\documentclass{article}

\usepackage[final]{neurips_data_2024}

\usepackage[utf8]{inputenc} % allow utf-8 input
\usepackage[T1]{fontenc}    % use 8-bit T1 fonts
\usepackage{hyperref}       % hyperlinks
\usepackage{url}            % simple URL typesetting
\usepackage{booktabs}       % professional-quality tables
\usepackage{amsfonts}       % blackboard math symbols
\usepackage{nicefrac}       % compact symbols for 1/2, etc.
\usepackage{microtype}      % microtypography
\usepackage{xcolor}         % colors

\usepackage{subcaption}
\usepackage{microtype}
\usepackage{glossaries} 

\newacronym{seb}{SEB}{Scandinavian Embedding Benchmark}
\newacronym{mteb}{MTEB}{Massive Text Embedding Benchmark}
\newacronym{rag}{RAG}{retrieval augmented generation}

\usepackage{multirow}
\usepackage{amssymb} % for checkmarks 
\usepackage{pifont} % for \xmark
\newcommand{\xmark}{\ding{55}}
\usepackage{xcolor}
\definecolor{darkergreen}{RGB}{0,120,0} % You can adjust the RGB values to get the desired shade
\definecolor{darkerred}{RGB}{160,0,0} % You can adjust the RGB values to get the desired shade
\newcommand{\greenplus}{{\color{darkergreen}\textbf{+}}}
\newcommand{\greencheck}{{\color{darkergreen}\textbf{\checkmark}}}
\newcommand{\redcross}{{\color{darkerred}\textbf{\xmark}}}

\usepackage{epigraph}

\usepackage{array} % for left aligned fixed length columns
\newcolumntype{L}[1]{>{\raggedright\let\newline\\\arraybackslash\hspace{0pt}}p{#1}}
\newcolumntype{H}{>{\setbox0=\hbox\bgroup}c<{\egroup}@{}}

\usepackage{tikz}
\newcommand*\circled[1]{\tikz[baseline=(char.base)]{
            \node[shape=circle,draw,inner sep=2pt] (char) {#1};}}

\title{The Scandinavian Embedding Benchmarks:\\ Comprehensive Assessment of Multilingual and Monolingual Text Embedding}

\author{Kenneth Enevoldsen \\
  Aarhus University \\ 
  \texttt{kenneth.enevoldsen@cas.au.dk} \\\And
  Márton Kardos \\
  Aarhus University\\ 
  \texttt{martonkardos@cas.au.dk} \\\AND
  Niklas Muennighoff  \\
  \texttt{n.muennighoff@gmail.com} \\\And
  Kristoffer Laigaard Nielbo \\
  Aarhus University \\
  \texttt{kln@cas.au.dk} \\}

\begin{document}

\maketitle

\begin{abstract}
    The evaluation of English text embeddings has transitioned from evaluating a handful of datasets to broad coverage across many tasks through benchmarks such as MTEB. However, this is not the case for multilingual text embeddings due to a lack of available benchmarks. To address this problem, we introduce the \gls{seb}. \glsfirst{seb} is a comprehensive framework that enables text embedding evaluation for Scandinavian languages across 24 tasks, 10 subtasks, and 4 task categories. Building on \glsfirst{seb}, we evaluate more than 26 models, uncovering significant performance disparities between public and commercial solutions not previously captured by MTEB. We open-source \glsfirst{seb}\footnote{\url{https://github.com/KennethEnevoldsen/scandinavian-embedding-benchmark}} and integrate it with MTEB, thus bridging the text embedding evaluation gap for Scandinavian languages.

\end{abstract}

\section{Introduction}
    
Natural language embeddings are used in a diverse range of applications, including clustering~\citep{liu2011survey, Angelov2020Top2VecDR}, 
text mining~\citep{jiang2015training}, 
semantic search~\citep{reimers2019sentence,muennighoff2022sgpt} 
and feature representation~\citep{alayrac2022flamingo}. 
Furthermore, embeddings are crucial in \gls{rag} systems \citep{borgeaud2022improving}, particularly for low- to mid-resource languages and domains. \gls{rag} systems enable the enrichment of generative models with the knowledge that might be underrepresented or absent during training. Thus, they can play a role in broadening linguistic and domain coverage.

% Applying generative language models or cross encoders is often infeasible due to the high computational overhead of pairwise comparison \citep{sentence_bert}.

With the breadth of applications for text embeddings, a proper evaluation of their quality is critical. Recent work has proposed \gls{mteb} \citep{muennighoff-etal-2023-mteb}, a benchmark for evaluating the quality of document embeddings for a wide variety of tasks. \gls{mteb} improves upon prior benchmarks by addressing the lack of evaluations across tasks. This has led to the widespread adoption of the benchmark for evaluating natural language embeddings.

However, while \gls{mteb} substantially improves the evaluation of text embeddings, the benchmark has the following shortcomings:

\begin{enumerate}
    \item \textbf{Support for non-English evaluation:} \gls{mteb} contains only limited support for evaluating non-English embeddings and multiple task categories are predominantly covered by translated datasets (classification) and important task such as retrieval has no multilingual support.
    \item \textbf{Reproducibilty:} \gls{mteb} does not include model implementations in the benchmark's code\footnote{This can, for instance, be seen in issues such as \url{https://github.com/embeddings-benchmark/mteb/issues/109}}. This is especially problematic since recent approaches such as prompt-based embedding models~\citep{muennighoff2022sgpt,xiao2023c,su2022one}, Matryoshka embeddings \citep{NEURIPS2022_c32319f4} introduce variables which can dramatically influence performance.
    \item \textbf{Coverage:} While \gls{mteb} has broad coverage across tasks, its domain coverage is still limited, as it primarily includes datasets from academic articles, social media, and web sources. This lack of coverage is especially pronounced for non-English tasks.
\end{enumerate}

Our work is driven by the reality that Scandinavian research, public institutions, and industry have to make decisions about their choice of text embedding model for various use cases.
These choices are currently made in the absence of a reliable standard to evaluate text embedding models' performance on Scandinavian languages.
As a result, these institutions have relied on proxies, such as models' performance on predominantly English benchmarks or Bitext mining tasks. 
As we demonstrate, performance on these tasks is not necessarily transferable to Scandinavian applications, thus not properly accounting for these institutions' requirements.
By introducing a benchmark tailored for Scandinavian languages, we aim to aid these organizations in making informed decisions.
Additionally, SEB will presumably support the development of Scandinavian embedding models by providing a standardized means for evaluating new models and comparing them against previously existing ones.

\subsection{Contributions}

% A lot of these datasets have never (to our knowledge) been used to evaluate embedding models, - external validity
To mitigate these issues, we present \gls{seb} a benchmark for embedding evaluation of the Mainland Scandinavian languages: Danish (da), Swedish (sv), and Norwegian (Bokmål (nb) and Nynorsk (nn)) as well as the Danish dialect Bornholmsk (da-bornholm). 
Due to the limited resources available for these languages we choose to utilize the substantial cross-lingual transfer between these languages demonstrated by \citeauthor{nielsen-2023-scandeval} (\citeyear{nielsen-2023-scandeval}); this supports collectively benchmarking the Mainland Scandinavian languages to broaden the coverage otherwise limited for these languages.
% A recent study by \citeauthor{nielsen-2023-scandeval} (\citeyear{nielsen-2023-scandeval}) shows substantial cross-lingual transfer among these three languages, suggesting that the mainland Scandinavian languages can meaningfully be benchmark together. should be used jointly for benchmarking language models..
\gls{seb} makes the following main contributions; \circled{1} it greatly expands the evaluation of embedding for Scandinavian to multiple tasks (see \autoref{tab:task_type_coverage}) as well as across a wide range of domains (see \autoref{tab:domain_coverage}); \circled{2} \gls{seb} implements a model registry that allows for the easy addition of new models as well as documents the exact implementation of existing models evaluated in the benchmark.
Lastly, \circled{3} \gls{seb} expands and extends \gls{mteb} by porting all tasks, allowing for the expansion of \gls{mteb} to a fully-fledged multilingual benchmark for embeddings. 
Using \gls{seb} we evaluate 26 representative models and APIs within this work and present additional models in an interactive online dashboard.\footnote{\url{https://kennethenevoldsen.github.io/scandinavian-embedding-benchmark/}}

% potentially cut:
% Furthermore, multiple of the datasets added have, to the authors' knowledge, never been used to evaluate document embedding models, thus providing an external validity of existing multilingual models.

%Our benchmark specifically targets bitext mining for local dialects  (Bornholm Parallel) or written forms (Norwegain Courts), languages learning (DaLAJ), governmental QA (SweFAQ)

% mini result?
% \textbf{We additionally show ...}

\begin{figure*}
    \centering
    \includegraphics[width=.9\linewidth]{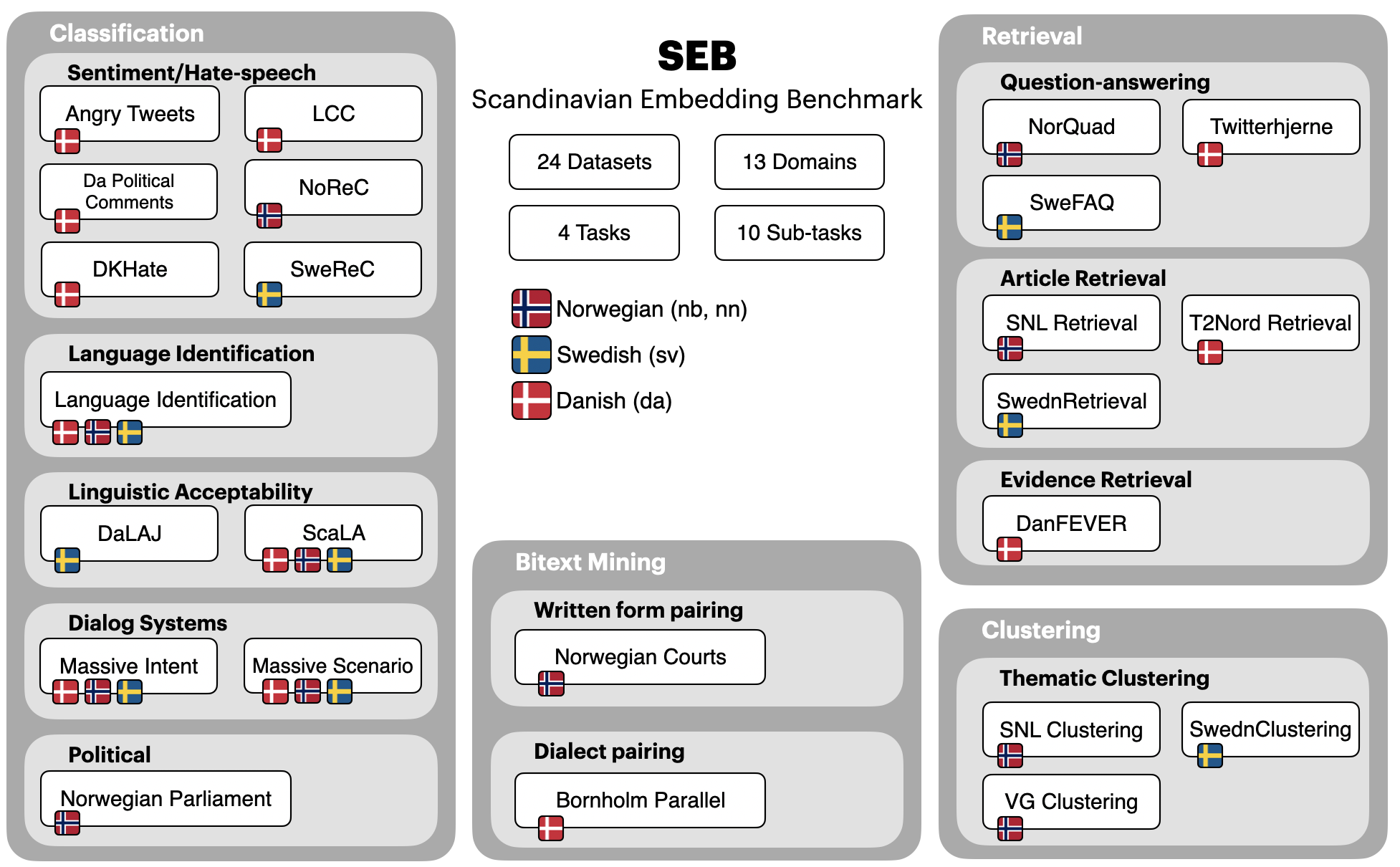}
    \caption{An overview of the tasks and datasets in \gls{seb}. Flags denote the languages of the datasets.}
    \label{fig:overview}
\end{figure*}

\section{Related Work}
    \subsection{Benchmarks}

Benchmarks are important tools for model development that enable the assessment of significant performance improvements.
Prior benchmarks for evaluating text embeddings focused on specific embedding qualities; BEIR \citep{Thakur2021BEIRAH} and MIRACL \citep{zhang2023miracl} assessed embedding efficacy in information retrieval across diverse domains or languages, while SentEval \citep{conneau-kiela-2018-senteval} integrated various SemEval datasets for sentence encoding evaluation using semantic text similarity (STS) tasks. \gls{mteb} \citep{muennighoff-etal-2023-mteb} amalgamated and expanded these methodologies to cover eight different tasks. While \gls{mteb} includes more than 112 languages, most of this linguistic variation originates from only a handful of tasks, notably bitext mining \citep{TatoebaCorpus} or translated datasets \citep{fitzgerald2022massive}. Scandinavian languages are only represented in two datasets for intent and scenario classification \citep{fitzgerald2022massive}, both of which are translations. Thus, the benchmark contains no naturally occurring text for either of these languages.

While benchmarks for Scandinavian languages have been developed, most -- akin to (Super)GLUE \citep{wang-etal-2018-glue, wang2019superglue} -- seek to evaluate the performance of multiple natural language understanding tasks. These include monolingual benchmarks such as the Swedish superlim \citep{berdicevskis-etal-2023-superlim}, the Norwegian NorBench \citep{samuel-etal-2023-norbench}, or cross-lingual benchmarks such as ScandEval \citep{nielsen-2023-scandeval}. While these benchmarks are instrumental for developing Scandinavian models, none focus on evaluating text embeddings for, e.g., retrieval or clustering.

\subsection{Text Embeddings}

Over time, the development of dense text embedding models has evolved from focusing on individual words~\citep{mikolov2013distributed, pennington2014glove} to encompass entire sentences~\citep{conneau2017supervised,ni2021sentence}, and currently extends to processing multiple sentences in a wide range of tasks~\citep{xiao2023c,su2022one,muennighoff2024generative}. As is common in natural language processing~\citep{xue2020mt5}, English-centric models have led this development, followed by multilingual models with only a short delay. While multilingual word embedding models already exist~\citep{artetxe2019massively}, multitask sentence-level multilingual embedding models are just beginning to emerge~\citep{chen2024bge,wang2022text}. However, their progress is hindered by the lack of comprehensive evaluation for multilingual tasks. This evaluation gap hinders progress in the field, preventing us from effectively evaluating model improvements. Our work aims to address this problem to enable further progress and proliferation of multilingual text embeddings.

% methods
\section{The Benchmark}
    
\subsection{Design and Curation Rationale}
\gls{seb} seeks to provide an estimate of the quality of embedding for Scandinavian languages and multilingual use cases. To do so, we focus on

\noindent
\textbf{a) Coverage:}
% or diversity
The benchmark should cover a wide variety of tasks spanning distinctly different domains, usages, and embedding tasks; \gls{seb} compromises 24 datasets spanning at least 12 domains across nine different tasks with broad coverage for each language.

\noindent
\textbf{b) Cultural integrity and model equity:}
Recent studies \citep{berdicevskis-etal-2023-superlim,nielsen-2023-scandeval,muennighoff-etal-2023-mteb} have increasingly adopted the strategy of leveraging translated English datasets as a means to evaluate the performance of models in low-resource language contexts. However, we avoid adding such translations, aiming to represent Scandinavian contexts accurately and mitigate the risk of artificially inflating multilingual model capabilities. 
This decision stems from the recognition that multilingual models, often trained on parallel or translated data \citep{reimers-gurevych-2020-making}, may exhibit inflated performance when evaluated on similarly translated tasks --- a hypothesis that, while plausible, remains to be conclusively shown.
We choose to keep the existing translated datasets from \gls{mteb} within \gls{seb} to maintain compatibility.

% we examine the relation between the existing trasnslated datasets with \gls{mteb} and the newly added dataset in \autoref{sec:similarity}.

\noindent
\textbf{c) Cross-lingual generalization:} 
Given the limited availability of datasets for the Scandinavian languages, we rely on the high degree of cross-lingual transfer \citep{nielsen-2023-scandeval} to estimate model performance more accurately. This approach capitalizes on intrinsic linguistic similarities and shared cultural contexts to bridge data gaps.

\noindent
\textbf{d) Reproducibility and Accessibility:} 
\gls{seb} expands upon the reproducibility of \gls{mteb} by including a model registry for all evaluated models to ensure the exact method (e.g., model prompts) for obtaining the results is known. Furthermore, to ensure that the benchmark is as widely accessible as possible, we have limited the size of most datasets to a maximum of 2048 examples. For most models, this allows running the benchmark on a consumer-grade laptop while ensuring proper performance estimation. The benchmark also implements a public cache, allowing users to experiment without needing to rerun models run by others.

In addition to these criteria, \gls{seb} follows the desiderata outlined by \citeauthor{muennighoff-etal-2023-mteb} (\citeyear{muennighoff-etal-2023-mteb}), allowing for easy extension of the benchmark and providing a simple API and command-line interface making it easy to benchmark models that are not part of SEB by default.

% Considerations:
% - A good model should encode both the language and the semantic (we argue that encoding the language is a part of encoding the semantics)

\subsection{Datasets}
\begin{figure*}
    \centering
    \includegraphics[width=0.68\linewidth]{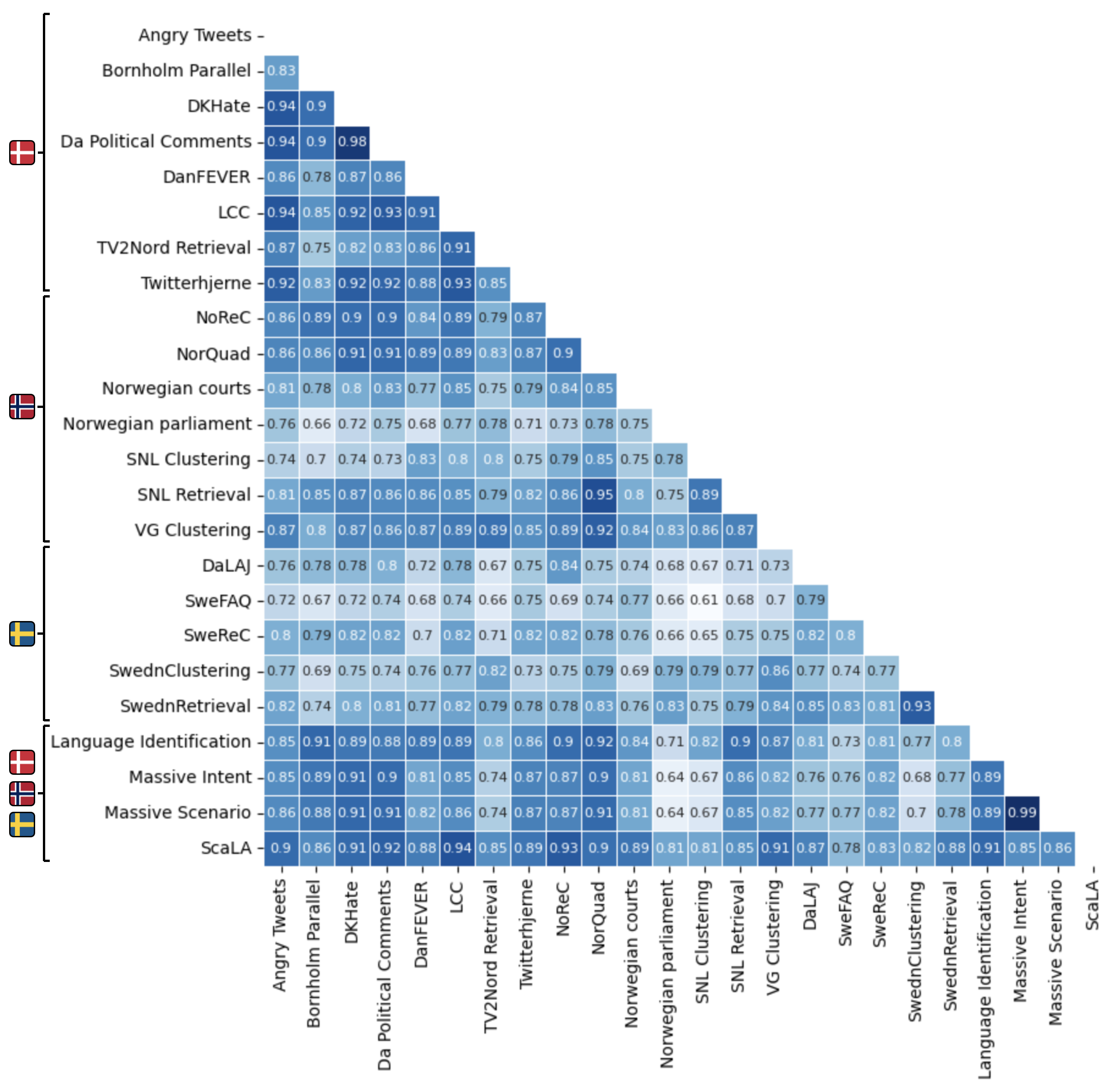}
    \caption{Dataset similarity between the datasets included within \gls{seb}. Embeddings are obtained by applying the embed-multilingual-v3.0 on 100 randomly sampled documents. Similarity is computed using cosine similarity.}
    \label{fig:dataset_similarity}
\end{figure*}

We present an overview of the tasks in \gls{seb} in \autoref{fig:overview}. Additionally, we have created an overview of the datasets in \autoref{tab:dataset_desc}, including dataset statistics and a short description of each dataset. \autoref{sec:app_evaluation} described the method of evaluation, and  \autoref{sec:app_dataset_construction} described the formalization of the specific datasets to the task.
\gls{seb} seeks to cover a large variety of domains and task types, greatly expanding upon what was previously available for non-English languages within \gls{mteb} (see \autoref{tab:domain_coverage} and \ref{tab:task_type_coverage}). To allow for the exploration, we add an embedding map of samples from the dataset in \autoref{sec:appendix_embedding_map}, where it is clearly seen that the datasets occupy different clusters. Similarly,  \autoref{fig:dataset_similarity} reveals distinctly different clusters of datasets, e.g., the high similarity between SNL Retrieval and NorQuad as both are constructed from encyclopedic sources while distinct datasets such as SweFAQ \citep{berdicevskis-etal-2023-superlim}, covering FAQ related to the public sector.

% Considerations:
% - A good model should encode both the language and the semantic (we argue that encoding the language is a part of encoding the semantics)

% Use for ACL ARR
% \input{tables/coverage_task_type}
% \input{tables/coverage_domain}

% Use for Neurips D&B

\begin{table}
\caption{Coverage on Mainland Scandinavian languages. The green plus (\greenplus) indicates newly added, while "\greenplus\greenplus" indicates previously not covered in MTEB by any language. The parenthesis is due to the LCC \cite{lcc} containing the domains, but only to a limited extent. Black checks (\checkmark) indicate domains covered in \gls{mteb} for Scandinavian Languages, though only within translated datasets. The domains follow the categorization of the Universal Dependencies \cite{nivre-etal-2017-universal}.}
\begin{subtable}{.45\linewidth}
\centering
\caption{Domain Coverage}
{\footnotesize
    \begin{tabular}{lcccc}
        \hline
         & \multicolumn{4}{c}{\textbf{Language}} \\ 
        \textbf{Domain} & da & nb & nn & sv \\ \hline 
        Academic & (\greenplus) & & & \\
        Bible &  & \\
        Blog &  & \\
        Fiction & \greenplus & \greenplus & \greenplus & \greenplus   \\
        Government & \greenplus\greenplus & \greenplus\greenplus & \greenplus\greenplus & \greenplus\greenplus \\
        Legal & (\greenplus\greenplus) & \greenplus\greenplus & \greenplus\greenplus &  \\
        Medical &  & & &  \\
        News & \greenplus & \greenplus & & \greenplus \\
        Non-fiction & \greenplus  & \greenplus & & \greenplus\\
        Poetry &  (\greenplus\greenplus) & & &  \\
        Reviews &  & \greenplus & &  \\
        Social & \greenplus & & & \greenplus \\
        Spoken & (\checkmark) & (\checkmark) & & (\checkmark) \\
        Wiki & \greenplus & \greenplus & \greenplus & \greenplus \\
        Web & \greenplus & &  & \greenplus\\ \hline
    \end{tabular}
}
\label{tab:domain_coverage}
\end{subtable}%
\begin{subtable}{.45\linewidth}
\centering
\caption{Task Coverage}
{\footnotesize
\begin{tabular}{lcccc}
\hline
                                         & \multicolumn{4}{c}{\textbf{Language}} \\
\textbf{Task}                             & da & nb & nn & sv  \\  \hline
\textbf{Retrieval}                       &            &            &            &            \\
\hspace{1mm} Question answering          & \greenplus & \greenplus &            & \greenplus \\
\vspace{1mm}\hspace{1mm} Article retrieval           & \greenplus\greenplus & \greenplus\greenplus &            & \greenplus\greenplus \\
\textbf{Bitext Mining}                   &            &            &            &            \\
\hspace{1mm} Dialect pairing             & \greenplus\greenplus & \greenplus\greenplus & \greenplus\greenplus & \greenplus\greenplus \\
\vspace{1mm}\hspace{1mm} Written form pairing             & \greenplus\greenplus & \greenplus\greenplus & \greenplus\greenplus & \greenplus\greenplus \\

\textbf{Classification}                  &            &            &            &            \\
\hspace{1mm} Political                   &            & \greenplus\greenplus & \greenplus\greenplus & \greenplus\greenplus \\
\hspace{1mm} Language Identification     & \greenplus & \greenplus\greenplus & \greenplus\greenplus & \greenplus\greenplus \\
\hspace{1mm} Linguistic Acceptability    & \greenplus\greenplus & \greenplus\greenplus & \greenplus\greenplus & \greenplus\greenplus \\
\hspace{1mm} Sentiment/Hate Speech       & \greenplus & \greenplus &            & \greenplus \\
\vspace{1mm}\hspace{1mm} Dialog Systems              & (\checkmark) & (\checkmark) & (\checkmark) & (\checkmark) \\
\textbf{Clustering}                      &            &            &            &            \\
\hspace{1mm} Thematic Clustering         & \greenplus & \greenplus &            & \greenplus \\ \hline
\end{tabular}
}
\label{tab:task_type_coverage}
\end{subtable}%

\end{table}

% Considerations:
% - A good task should be saturated
% Some datasets lend themselves to the formalized in different ways. For instance, a Summerization dataset could be formalized as an STS task between summaries and articles, while similarly being formulated as a retrieval task. In table X we examine the difference between the two formalizations and show that task N

% - A good model should encode both the language and the semantic (we argue that encoding the language is a part of encoding the semantics)

% Size of datasets. Should be small enough, while maintaining consistent scores. If a test set is available we typically use that one or a subsampled version of it. 

\section{Methodology}

% H = hidden column (makes editing the table easier)

\begin{table*}
    \centering

    \caption{
    Performance across task-type categories and languages in \gls{seb}. 
    The best score in each model category is highlighted in bold. Additional model evaluation can be found on the public Dashboard. 
    Rank is calculated across all models within the benchmark. 
    The brackets indicate the 95\% confidence interval, obtained by bootstrapping 100 repetitions with tasks to minimize the impact of any single task. 
    The symbol "*" signifies when the top-performing model significantly outperforms the second-best model within the same category at a 0.05 significance threshold. 
    Ranks are reported using two significant figures.
    }
    
    {\footnotesize
    \begin{tabular}{lH|cc|cccc|cccc}
        \toprule
                               &  & &  &   \multicolumn{4}{c|}{\textbf{Task-Type}} &  \multicolumn{4}{c}{\textbf{Language}}\\
          & Public & Avg. rank &  Avg. & Bitext & Class. & Clust. & Retr.  & da & nb & nn & sv \\
         \midrule
         Num. Datasets ($\rightarrow$) & & 24 & 24 &  2 & 12 & 3 & 7 & 12 & 11 & 3 & 9 \\ 
         \midrule
         \midrule
         \textit{Self-Supervised Models} &  \\
         \midrule
            dfm-encoder-large & \greencheck & 23 (19-26) & 41.4 & 46.8 & 56.5 & 26.9 & 20.1 & 47.7 & 47.4 & 72.5 & 43.7 \\
            \quad + SimCSE & \greencheck & 19 (16-22) & \textbf{46.6} & 50.9 & 58.4 & 26.9 & \textbf{33.7} & \textbf{52.2} & 51.3 & 74.3 & 42.0 \\
            xlm-roberta-large & \greencheck & 25 (21-30) & 35.3 & 19.1 & 54.6 & 28.1 & 10.0 & 39.6 & 41.3 & 58.0 & 44.5 \\
            nb-bert-large & \greencheck & \textbf{17} (13-20) & 46.0 & 47.3 & \textbf{59.3} & \textbf{35.7} & 27.3 & 46.8 & \textbf{57.2} & \textbf{80.4} & \textbf{50.2} \\
            nb-bert-base & \greencheck & 21 (18-25) & 42.1 & \textbf{51.0} & 57.0 & 31.8 & 18.4 & 43.6 & 53.0 & 79.2 & 47.7 \\
            bert-base-swedish & \greencheck & 28 (24-31) & 35.2 & 39.1 & 49.7 & 26.2 & 13.2 & 34.0 & 41.1 & 62.2 & 43.6 \\
            fasttext-cc-da & \greencheck & 32 (29-34) & 37.3 & 42.4 & 48.8 & 21.8 & 22.7 & 39.0 & 43.2 & 66.4 & 38.7  \\
            fasttext-cc-nn & \greencheck & 32 (29-35) & 35.8 & 47.6 & 46.2 & 22.1 & 20.4 & 34.6 & 43.9 & 69.1 & 37.1 \\
            fasttext-cc-nb & \greencheck & 30.0 (27-32) & 37.5 & 43.2 & 48.7 & 24.2 & 22.2 & 37.5 & 45.6 & 67.7 & 38.9\\
            fasttext-cc-sv & \greencheck & 31.5 (29-34) & 36.0 & 43.3 & 47.3 & 22.0 & 20.4 & 34.9 & 41.3 & 63.4 & 40.6\\
        
        \midrule
        \textit{Supervised Models} & \\
        \midrule
            multilingual-MiniLM-L12 & \greencheck & 20 (17-23) & 50.0 & 51.0 & 53.7 & 31.7 & 51.1 & 49.9 & 52.7 & 58.3 & 50.3\\
            multilingual-mpnet-base & \greencheck & 16 (13-20) & 53.2 & 52.7 & 56.5 & 32.7 & 56.5 & 53.0 & 55.8 & 59.6 & 53.3\\
            labSE & \greencheck & 18 (15-21) & 50.5 & 69.1 & 53.6 & 29.0 & 48.9 & 50.9 & 52.9 & 59.4 & 48.7\\
           sentence-bert-swedish & \greencheck & 23 (19-26) & 46.6 & 43.3 & 51.0 & 35.6 & 44.6 & 43.2 & 48.2 & 62.7 & 54.7\\
            e5-mistral-7b-instruct  & \greencheck & \textbf{8.7} (6.8-12) & 60.4 & \textbf{70.8} & 61.7 & 35.7 & 66.0 & \textbf{61.7} & 62.9 & 68.8 & 60.4 \\
            multilingual-e5-large & \greencheck & 8.8 (6.0-12) & \textbf{60.7} & 60.1 & \textbf{62.5} & 34.2 & \textbf{69.1} & 61.1 & \textbf{63.1} & \textbf{73.9} & \textbf{62.8} \\
            multilingual-e5-base & \greencheck & 12 (9.4-15) & 57.9 & 61.4 & 60.1 & 34.0 & 63.5 & 58.6 & 60.9 & 72.0 & 58.5 \\
            multilingual-e5-small & \greencheck & 14 (11-16) & 56.4 & 61.6 & 58.1 & \textbf{36.9} & 60.3 & 56.5 & 58.9 & 69.5 & 57.1 \\
            translate-e5-large & \greencheck & 21 (18-24) & 47.7 & 50.7 & 54.7 & 27.3 & 43.4 & 49.0 & 50.1 & 59.2 & 59.2 \\
            sonar-dan & \greencheck &  23 (20-26) & 43.4 & 70.5 & 53.5 & 19.6 & 28.6 & 48.3 & 46.0 & 63.7 & 42.9 \\
            sonar-nob & \greencheck & 25 (21-28) & 41.5 & 63.2 & 52.9 & 18.5 & 25.6 & 45.2 & 45.9 & 64.7 & 42.4 \\
            sonar-nno & \greencheck & 25 (22-28) & 41.5 & 65.5 & 52.8 & 17.3 & 25.7 & 45.5 & 45.1 & 63.2 & 42.6 \\
            sonar-swe & \greencheck & 24 (21-27) & 42.8 & 70.7 & 52.9 & 19.4 & 27.6 & 47.1 & 45.4 & 63.1 & 42.9 \\
        \midrule
        \textit{Embedding APIs} & \\
        \midrule
            text-embedding-3-large & \redcross & \textbf{5.8} (3.3-8.2) & \textbf{65.0} & \textbf{68.8} & 63.5 & 38.7 & \textbf{77.9} & \textbf{63.7} & \textbf{69.0} & \textbf{74.7} & \textbf{65.5}\\ 
            text-embedding-3-small & \redcross & 9.4 (7.7-12) & 61.0 & 66.7 & 59.7 & 38.3 & 71.3 & 59.7 & 64.7 & 70.2 & 60.4\\
            embed-multilingual-v3.0 & \redcross & 6.1 (3.8-8.9) & 64.1 & 64.2 & \textbf{63.6} & \textbf{40.2} & 75.2 & 62.6 & 68.5 & 74.1 & 64.3\\
         \toprule
    \end{tabular}
    }
    \label{tab:result_across_task_type}
\end{table*}

We describe the construction of the datasets in \autoref{sec:app_dataset_construction}. To keep our benchmark compatible with \gls{mteb} we follow a similar approach for computing scores, these are described in \autoref{sec:app_evaluation}.

\subsection{Models}
For our benchmarked models, we have chosen a series of representative models seeking to cover a range of model architectures, model sizes, and commercial APIs, as well as models claiming state-of-the-art results on various embedding tasks.
In addition, the online dashboard includes additional models not represented here. We group the models into self-supervised and supervised methods.

\noindent
\textbf{Self-supervised methods:} 

\textbf{Encoders} such as BERT models \citep{devlin-etal-2019-bert} including monolingual or Scandinavian models trained for Danish \citep{enevoldsen2023danish}, Norwegian \citep{kummervold-etal-2021-operationalizing} and Swedish \citep{rekathati2021introducing} as well as the multilingual model XLM-R \citep{conneau-etal-2020-unsupervised}. We also include a SimCSE \citep{gao-etal-2021-simcse} version of the dfm-encoder-large to indicate the potential performance gain by self-supervised pre-training. This model is trained on sentences extracted from the Danish Gigaword \citep{stromberg-derczynski-etal-2021-danish} using default parameters\footnote{For exact specification see the model card; \url{https://huggingface.co/KennethEnevoldsen/dfm-sentence-encoder-large}}.

\noindent
As a candidate for \textbf{Static Word Vectors}, we include four fastText \citep{joulin2016fasttext, joulin2017bag, bojanowski2017enriching} models for Danish, Swedish, and Norwegian Bokmål and Nynorsk respectively.

\noindent 
 \textbf{Supervised Methods:}

For \textbf{encoders}, we benchmark LaBSE \citep{feng-etal-2022-language}, which is based on BERT but further pre-trained on a parallel corpus. Further, we evaluate the multilingual MiniLM models and MPNet models \citep{reimers-gurevych-2019-sentence, song2020mpnet, wang-etal-2021-minilmv2}, which are trained on diverse datasets. We also include the SONAR models \citep{duquenne2023sonar} as they claim improved performance over LabSE. In addition, we include the Swedish sentence transformers~\citep{rekathati2021introducing} trained with knowledge distillation from an English model~\citep{reimers-gurevych-2020-making}.

\noindent
Because the development of Scandinavian \textbf{decoders} is only in its early stages \citep{enevoldsen2023danish, ekgren-etal-2022-lessons}, we utilize the e5-mistral model \citep{wang2022text,wang2023improving} as it presents a competitive model in the category.

\noindent
\textbf{Commercial embedding APIs:} We additionally include the embedding APIs of Cohere \footnote{\url{https://txt.cohere.com/introducing-embed-v3/}} and OpenAI \footnote{\url{https://openai.com/blog/new-embedding-models-and-api-updates}} to compare openly available models with commercial solutions.

\noindent
Lastly, we add \textbf{Translate and embed} as a baseline model for comparing naïvely translating to English and then embedding with high-quality English models. To allow for comparison with multilingual models, we include both the large English e5 model and all sizes of its multilingual variants \citep{wang2022text}. We use the multilingual M2M100 model \citep{fan2020englishcentric} for the translation. For translation, we assume the language is known. This avoids accumulating errors due to language detection, and in many applications, the language would be known. We assume Danish as the origin for tasks requiring multiple languages, such as bitext mining.
\section{Results}
    
% general + supervised vs apis
In \autoref{tab:result_across_task_type}, we see that the best-performing model is either of the commercial APIs of OpenAI and Cohere followed by the publicly available multilingual e5 model series \citep{wang2022text}. This stands in contrast to developments observed from ScandEval \citep{nielsen-2023-scandeval}, where notably smaller monolingual or Scandinavian models have proven to be competitive, often surpassing significantly larger multilingual models.
% supervised vs unsupervised
Similar to \gls{mteb} \citep{muennighoff-etal-2023-mteb}, we find a pronounced performance between self-supervised methods and their supervised counterparts, although we see that notable gains can be obtained from unsupervised pre-training \citep{gao-etal-2021-simcse}. In general, however, utilizing unsupervised contrastive pretraining pales in comparison to popular multilingual models of smaller size.

In \autoref{tab:result_across_domain}, we see the performance across domains. Generally, we see that model rankings remain relatively stable across these domains, with the e5 models \citep{wang2022text} and the commercial APIs taking a consistent lead. However, we also see that in domains such as the legal domain, spoken language, and fiction, we see the e5-mistral-7b-instruct outcompeting commercial solutions.

If we examine individual subtask (see \autoref{sec:result_pr_task}) Pretrained encoders perform surprisingly well on language acceptability and language detection tasks. This is likely due to a trade-off between semantics and syntax. Self-supervised training on natural language will likely assign significance to syntactic nuances, while models trained on semantic tasks ignore some syntactical errors favoring semantics.

\noindent
\textbf{Performance across task-types:}
Models that have been contrastively trained on sentence pairs or finetuned for a set of common tasks typically outperform pre-trained models, especially in retrieval contexts, while LaBSE \citep{feng-etal-2022-language} and the SONAR models \citep{duquenne2023sonar}, which has been designed for bitext-mining purposes, excels at the task.

The largest gap between commercial and public models is in retrieval, where performance drops more than eight points. While notable improvements have been achieved in publicly available embedding models for English retrieval tasks \citep{wang2023improving}, similar results are yet to be achieved in multilingual contexts.
Bitext mining is the only category in which open solutions outperform commercial solutions.

\noindent
\textbf{Translate then embed:}
When comparing the 'translate-then-embed' model against the multilingual e5 models, we see that in almost all cases, the multilingual models perform better even when comparing across size categories. While performance could likely be improved by utilizing state-of-the-art embedding and translation models, we see few benefits to this approach due to increased computational costs, model complexity, and competitive approaches for knowledge distillation across languages \citep{reimers-gurevych-2020-making}.

% We should likely discuss whether this is an issue or not

% A correlation matrix between task types would be really nice

% One thing we should somehow measure, is how well models perform for Scandinavian-specific knowledge as opposed to something generic, like Wikipedia.

% \subsection{Comparison with other benchmarks}
% can the SEB be used to rank models (if so it is notably faster than ScandEval) and could be used for monitoring and model development. does including sentence transformers skews the correlation?

% crosslingual transfer on monolingual models?

% \section{Rank comparison with MTEB}
% either compare ranking of multilingual models 
    % correlation multilingual models

\subsection{Efficiency}
\begin{figure*}
    \centering
    \includegraphics[width=0.78\linewidth]{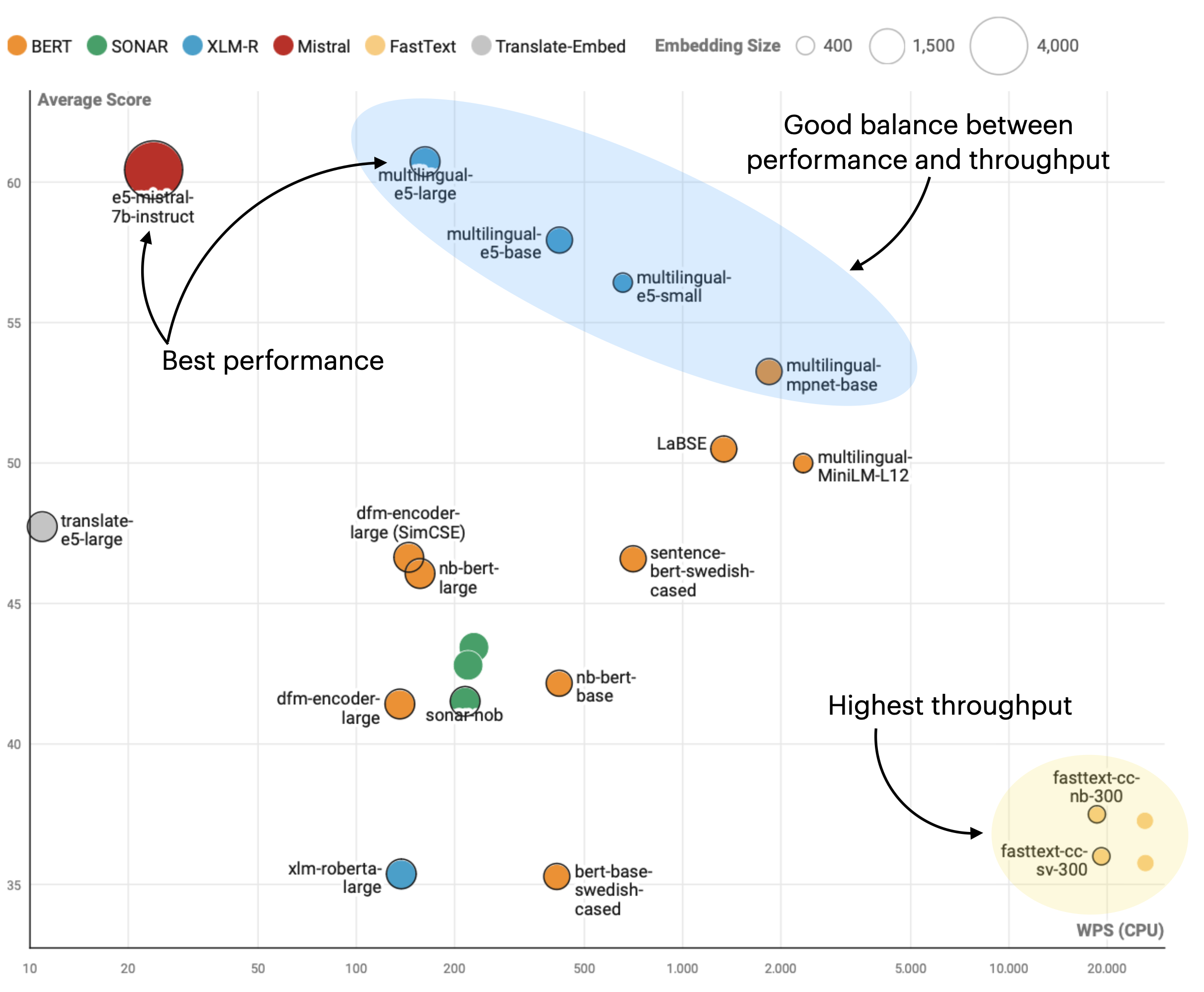}
    \caption{Performance and speed of embeddings models. The size of the circles denotes the embedding size, and the color denotes the model type. Note that commercial APIs are not included. WPS stands for words per second. We avoid highlighting all models to improve readability.}
    \label{fig:performance_x_speed}
\end{figure*}

We examine the trade-offs between performance and speed in \autoref{fig:performance_x_speed}. Speed was benchmarked on Dell PowerEdge C6420 Intel(R) Xeon(R) Gold 6130 CPUs with 32 cores/CPU. We see the following categories of relevance;

\noindent
\textbf{Highest Throughput} 
FastText models offer the highest throughput while maintaining an average performance exceeding that of the multilingual XLM-R \citep{conneau-etal-2020-unsupervised}.

\noindent
\textbf{Maximum Performance}
Achieving optimal performance is possible with the multilingual-e5-large or the e5-mistral-7b-instruct, which rivals the smaller commercial embedding APIs.

\noindent
\textbf{Balanced Performance:}
The best balance between performance, throughput, and embedding size is seen in the multilingual e5 models series, which performs competitively on the benchmark. The multilingual-mpnet-base also offers a competitive balance between throughput and performance, despite its larger embedding size.

% \section{Discussion}
% potentially add:
% reiterate on the Necessity of SEB
% discuss Performance difference between publicly available vs. commercial models
% Future perspectives

\subsection{Limitations and Future Perspectives}
    
\textbf{Domain Coverage}: Despite the advancements introduced by \gls{seb}, the benchmark could further benefit from encompassing domains crucial to the welfare states of Scandinavia, such as legal, governmental, and medical fields, which are currently only partly covered or unaddressed. Current tasks predominantly feature non-fiction literature, such as encyclopedias and news, yet the rising interest in digital humanities \citep{su2020digital} suggests the inclusion of fiction, poetry, historical texts, and religious documents in future updates could be valuable. Additionally, the benchmark currently lacks some task categories, such as pair classification and document reranking.

\noindent
\textbf{Future Directions:} While this work announces the release of \gls{seb}, we plan to continually expand upon the benchmark. As this work continues to develop, we invite researchers to join us in expanding the evaluation of embedding models across a broad range of languages.

\section{Conclusion}
    
In this work, we introduced \glsfirst{seb}, a framework that addresses the evaluation gap for the mainland Scandinavian languages. \gls{seb} encompasses 24 tasks covering ten subtasks in four task categories and spanning mainland Scandinavian languages. 

We evaluate more than 50 models on \gls{seb} and show that there is still a notable gap in performance between publicly available text embedding models and their commercial counterparts, especially in retrieval contexts, as well as between monolingual and multilingual models. These findings highlight critical areas for future advancements. By open-sourcing \gls{seb} and integrating it with \gls{mteb}, we aim to encourage the development of robust Scandinavian and multilingual embedding models, inviting the research community to contribute to this evolving landscape.

\begin{ack}
    Part of the computation for this project was performed on the DeiC interactive HPC system via UCloud managed by the Danish consortium for \href{https://interactivehpc.dk/}{Interactive HPC}. The computation was performed using a 40GB NVIDIA Tesla A100 GPU. 
\end{ack}

\bibliography{anthology,custom}

\begin{thebibliography}{62}
\expandafter\ifx\csname natexlab\endcsname\relax\def\natexlab#1{#1}\fi

\bibitem[{Alayrac et~al.(2022)Alayrac, Donahue, Luc, Miech, Barr, Hasson, Lenc, Mensch, Millican, Reynolds et~al.}]{alayrac2022flamingo}
Jean-Baptiste Alayrac, Jeff Donahue, Pauline Luc, Antoine Miech, Iain Barr, Yana Hasson, Karel Lenc, Arthur Mensch, Katherine Millican, Malcolm Reynolds, et~al. 2022.
\newblock Flamingo: a visual language model for few-shot learning.
\newblock \emph{Advances in Neural Information Processing Systems}, 35:23716--23736.

\bibitem[{Angelov(2020)}]{Angelov2020Top2VecDR}
Dimitar Angelov. 2020.
\newblock \href {https://api.semanticscholar.org/CorpusID:221246303} {Top2vec: Distributed representations of topics}.
\newblock \emph{ArXiv}, abs/2008.09470.

\bibitem[{Artetxe and Schwenk(2019)}]{artetxe2019massively}
Mikel Artetxe and Holger Schwenk. 2019.
\newblock Massively multilingual sentence embeddings for zero-shot cross-lingual transfer and beyond.
\newblock \emph{Transactions of the Association for Computational Linguistics}, 7:597--610.

\bibitem[{Berdicevskis et~al.(2023)Berdicevskis, Bouma, Kurtz, Morger, {\"O}hman, Adesam, Borin, Dann{\'e}lls, Forsberg, Isbister, Lindahl, Malmsten, Rekathati, Sahlgren, Volodina, B{\"o}rjeson, Hengchen, and Tahmasebi}]{berdicevskis-etal-2023-superlim}
Aleksandrs Berdicevskis, Gerlof Bouma, Robin Kurtz, Felix Morger, Joey {\"O}hman, Yvonne Adesam, Lars Borin, Dana Dann{\'e}lls, Markus Forsberg, Tim Isbister, Anna Lindahl, Martin Malmsten, Faton Rekathati, Magnus Sahlgren, Elena Volodina, Love B{\"o}rjeson, Simon Hengchen, and Nina Tahmasebi. 2023.
\newblock \href {https://doi.org/10.18653/v1/2023.emnlp-main.506} {Superlim: A {S}wedish language understanding evaluation benchmark}.
\newblock In \emph{Proceedings of the 2023 Conference on Empirical Methods in Natural Language Processing}, pages 8137--8153, Singapore. Association for Computational Linguistics.

\bibitem[{Bojanowski et~al.(2017)Bojanowski, Grave, Joulin, and Mikolov}]{bojanowski2017enriching}
Piotr Bojanowski, Edouard Grave, Armand Joulin, and Tomas Mikolov. 2017.
\newblock Enriching word vectors with subword information.
\newblock \emph{Transactions of the Association for Computational Linguistics}, 5:135--146.

\bibitem[{Borgeaud et~al.(2022)Borgeaud, Mensch, Hoffmann, Cai, Rutherford, Millican, Van Den~Driessche, Lespiau, Damoc, Clark et~al.}]{borgeaud2022improving}
Sebastian Borgeaud, Arthur Mensch, Jordan Hoffmann, Trevor Cai, Eliza Rutherford, Katie Millican, George~Bm Van Den~Driessche, Jean-Baptiste Lespiau, Bogdan Damoc, Aidan Clark, et~al. 2022.
\newblock Improving language models by retrieving from trillions of tokens.
\newblock In \emph{International conference on machine learning}, pages 2206--2240. PMLR.

\bibitem[{Chen et~al.(2024)Chen, Xiao, Zhang, Luo, Lian, and Liu}]{chen2024bge}
Jianlv Chen, Shitao Xiao, Peitian Zhang, Kun Luo, Defu Lian, and Zheng Liu. 2024.
\newblock M3-embedding: Multi-lingual, multi-functionality, multi-granularity text embeddings through self-knowledge distillation.
\newblock Under review at ACL ARR.

\bibitem[{Conneau et~al.(2020)Conneau, Khandelwal, Goyal, Chaudhary, Wenzek, Guzm{\'a}n, Grave, Ott, Zettlemoyer, and Stoyanov}]{conneau-etal-2020-unsupervised}
Alexis Conneau, Kartikay Khandelwal, Naman Goyal, Vishrav Chaudhary, Guillaume Wenzek, Francisco Guzm{\'a}n, Edouard Grave, Myle Ott, Luke Zettlemoyer, and Veselin Stoyanov. 2020.
\newblock \href {https://doi.org/10.18653/v1/2020.acl-main.747} {Unsupervised cross-lingual representation learning at scale}.
\newblock In \emph{Proceedings of the 58th Annual Meeting of the Association for Computational Linguistics}, pages 8440--8451, Online. Association for Computational Linguistics.

\bibitem[{Conneau and Kiela(2018)}]{conneau-kiela-2018-senteval}
Alexis Conneau and Douwe Kiela. 2018.
\newblock \href {https://aclanthology.org/L18-1269} {{S}ent{E}val: An evaluation toolkit for universal sentence representations}.
\newblock In \emph{Proceedings of the Eleventh International Conference on Language Resources and Evaluation ({LREC} 2018)}, Miyazaki, Japan. European Language Resources Association (ELRA).

\bibitem[{Conneau et~al.(2017)Conneau, Kiela, Schwenk, Barrault, and Bordes}]{conneau2017supervised}
Alexis Conneau, Douwe Kiela, Holger Schwenk, Lo{\"\i}c Barrault, and Antoine Bordes. 2017.
\newblock \href {https://doi.org/10.18653/v1/D17-1070} {Supervised learning of universal sentence representations from natural language inference data}.
\newblock In \emph{Proceedings of the 2017 Conference on Empirical Methods in Natural Language Processing}, pages 670--680, Copenhagen, Denmark. Association for Computational Linguistics.

\bibitem[{Derczynski and Kjeldsen(2019)}]{derczynski-kjeldsen-2019-bornholmsk}
Leon Derczynski and Alex~Speed Kjeldsen. 2019.
\newblock \href {https://aclanthology.org/W19-6138} {Bornholmsk natural language processing: Resources and tools}.
\newblock In \emph{Proceedings of the 22nd Nordic Conference on Computational Linguistics}, pages 338--344, Turku, Finland. Link{\"o}ping University Electronic Press.

\bibitem[{Devlin et~al.(2019)Devlin, Chang, Lee, and Toutanova}]{devlin-etal-2019-bert}
Jacob Devlin, Ming-Wei Chang, Kenton Lee, and Kristina Toutanova. 2019.
\newblock \href {https://doi.org/10.18653/v1/N19-1423} {{BERT}: Pre-training of deep bidirectional transformers for language understanding}.
\newblock In \emph{Proceedings of the 2019 Conference of the North {A}merican Chapter of the Association for Computational Linguistics: Human Language Technologies, Volume 1 (Long and Short Papers)}, pages 4171--4186, Minneapolis, Minnesota. Association for Computational Linguistics.

\bibitem[{Duquenne et~al.(2023)Duquenne, Schwenk, and Sagot}]{duquenne2023sonar}
Paul-Ambroise Duquenne, Holger Schwenk, and Benoit Sagot. 2023.
\newblock Sonar: sentence-level multimodal and language-agnostic representations.
\newblock \emph{arXiv e-prints}.

\bibitem[{Ekgren et~al.(2022)Ekgren, Cuba~Gyllensten, Gogoulou, Heiman, Verlinden, {\"O}hman, Carlsson, and Sahlgren}]{ekgren-etal-2022-lessons}
Ariel Ekgren, Amaru Cuba~Gyllensten, Evangelia Gogoulou, Alice Heiman, Severine Verlinden, Joey {\"O}hman, Fredrik Carlsson, and Magnus Sahlgren. 2022.
\newblock \href {https://aclanthology.org/2022.lrec-1.376} {Lessons learned from {GPT}-{SW}3: Building the first large-scale generative language model for {S}wedish}.
\newblock In \emph{Proceedings of the Thirteenth Language Resources and Evaluation Conference}, pages 3509--3518, Marseille, France. European Language Resources Association.

\bibitem[{Enevoldsen et~al.(2023)Enevoldsen, Hansen, Nielsen, Egeb{\ae}k, Holm, Nielsen, Bernstorff, Larsen, J{\o}rgensen, H{\o}jmark-Bertelsen et~al.}]{enevoldsen2023danish}
Kenneth Enevoldsen, Lasse Hansen, Dan~S Nielsen, Rasmus~AF Egeb{\ae}k, S{\o}ren~V Holm, Martin~C Nielsen, Martin Bernstorff, Rasmus Larsen, Peter~B J{\o}rgensen, Malte H{\o}jmark-Bertelsen, et~al. 2023.
\newblock Danish foundation models.
\newblock \emph{arXiv preprint arXiv:2311.07264}.

\bibitem[{Fan et~al.(2020)Fan, Bhosale, Schwenk, Ma, El-Kishky, Goyal, Baines, Celebi, Wenzek, Chaudhary, Goyal, Birch, Liptchinsky, Edunov, Grave, Auli, and Joulin}]{fan2020englishcentric}
Angela Fan, Shruti Bhosale, Holger Schwenk, Zhiyi Ma, Ahmed El-Kishky, Siddharth Goyal, Mandeep Baines, Onur Celebi, Guillaume Wenzek, Vishrav Chaudhary, Naman Goyal, Tom Birch, Vitaliy Liptchinsky, Sergey Edunov, Edouard Grave, Michael Auli, and Armand Joulin. 2020.
\newblock \href {http://arxiv.org/abs/2010.11125} {Beyond english-centric multilingual machine translation}.

\bibitem[{Feng et~al.(2022)Feng, Yang, Cer, Arivazhagan, and Wang}]{feng-etal-2022-language}
Fangxiaoyu Feng, Yinfei Yang, Daniel Cer, Naveen Arivazhagan, and Wei Wang. 2022.
\newblock \href {https://doi.org/10.18653/v1/2022.acl-long.62} {Language-agnostic {BERT} sentence embedding}.
\newblock In \emph{Proceedings of the 60th Annual Meeting of the Association for Computational Linguistics (Volume 1: Long Papers)}, pages 878--891, Dublin, Ireland. Association for Computational Linguistics.

\bibitem[{FitzGerald et~al.(2023)FitzGerald, Hench, Peris, Mackie, Rottmann, Sanchez, Nash, Urbach, Kakarala, Singh, Ranganath, Crist, Britan, Leeuwis, Tur, and Natarajan}]{fitzgerald2022massive}
Jack FitzGerald, Christopher Hench, Charith Peris, Scott Mackie, Kay Rottmann, Ana Sanchez, Aaron Nash, Liam Urbach, Vishesh Kakarala, Richa Singh, Swetha Ranganath, Laurie Crist, Misha Britan, Wouter Leeuwis, Gokhan Tur, and Prem Natarajan. 2023.
\newblock \href {https://doi.org/10.18653/v1/2023.acl-long.235} {{MASSIVE}: A 1{M}-example multilingual natural language understanding dataset with 51 typologically-diverse languages}.
\newblock In \emph{Proceedings of the 61st Annual Meeting of the Association for Computational Linguistics (Volume 1: Long Papers)}, pages 4277--4302, Toronto, Canada. Association for Computational Linguistics.

\bibitem[{Gao et~al.(2021)Gao, Yao, and Chen}]{gao-etal-2021-simcse}
Tianyu Gao, Xingcheng Yao, and Danqi Chen. 2021.
\newblock \href {https://doi.org/10.18653/v1/2021.emnlp-main.552} {{S}im{CSE}: Simple contrastive learning of sentence embeddings}.
\newblock In \emph{Proceedings of the 2021 Conference on Empirical Methods in Natural Language Processing}, pages 6894--6910, Online and Punta Cana, Dominican Republic. Association for Computational Linguistics.

\bibitem[{Haas and Derczynski(2021)}]{haas-derczynski-2021-discriminating}
Ren{\'e} Haas and Leon Derczynski. 2021.
\newblock \href {https://aclanthology.org/2021.vardial-1.8} {Discriminating between similar {N}ordic languages}.
\newblock In \emph{Proceedings of the Eighth Workshop on NLP for Similar Languages, Varieties and Dialects}, pages 67--75, Kiyv, Ukraine. Association for Computational Linguistics.

\bibitem[{Holm(2024)}]{holm2024gllms}
S{\o}ren~Vejlgaard Holm. 2024.
\newblock Are gllms danoliterate? benchmarking generative nlp in danish.

\bibitem[{Ivanova et~al.(2023)Ivanova, Andreassen, Jentoft, Wold, and {\O}vrelid}]{ivanova-etal-2023-norquad}
Sardana Ivanova, Fredrik Andreassen, Matias Jentoft, Sondre Wold, and Lilja {\O}vrelid. 2023.
\newblock \href {https://aclanthology.org/2023.nodalida-1.17} {{N}or{Q}u{AD}: {N}orwegian question answering dataset}.
\newblock In \emph{Proceedings of the 24th Nordic Conference on Computational Linguistics (NoDaLiDa)}, pages 159--168, T{\'o}rshavn, Faroe Islands. University of Tartu Library.

\bibitem[{Jiang et~al.(2015)Jiang, Li, Huang, and Jin}]{jiang2015training}
Zhenchao Jiang, Lishuang Li, Degen Huang, and Liuke Jin. 2015.
\newblock Training word embeddings for deep learning in biomedical text mining tasks.
\newblock In \emph{2015 IEEE international conference on bioinformatics and biomedicine (BIBM)}, pages 625--628. IEEE.

\bibitem[{Joulin et~al.(2016)Joulin, Grave, Bojanowski, Douze, J{\'e}gou, and Mikolov}]{joulin2016fasttext}
Armand Joulin, Edouard Grave, Piotr Bojanowski, Matthijs Douze, Herv{\'e} J{\'e}gou, and Tomas Mikolov. 2016.
\newblock \href {https://api.semanticscholar.org/CorpusID:16196524} {Fasttext.zip: Compressing text classification models}.
\newblock \emph{ArXiv}, abs/1612.03651.

\bibitem[{Joulin et~al.(2017)Joulin, Grave, Bojanowski, and Mikolov}]{joulin2017bag}
Armand Joulin, Edouard Grave, Piotr Bojanowski, and Tomas Mikolov. 2017.
\newblock Bag of tricks for efficient text classification.
\newblock In \emph{Proceedings of the 15th Conference of the European Chapter of the Association for Computational Linguistics: Volume 2, Short Papers}, pages 427--431. Association for Computational Linguistics.

\bibitem[{Kummervold et~al.(2021)Kummervold, De~la Rosa, Wetjen, and Brygfjeld}]{kummervold-etal-2021-operationalizing}
Per~E Kummervold, Javier De~la Rosa, Freddy Wetjen, and Svein~Arne Brygfjeld. 2021.
\newblock \href {https://aclanthology.org/2021.nodalida-main.3} {Operationalizing a national digital library: The case for a {N}orwegian transformer model}.
\newblock In \emph{Proceedings of the 23rd Nordic Conference on Computational Linguistics (NoDaLiDa)}, pages 20--29, Reykjavik, Iceland (Online). Link{\"o}ping University Electronic Press, Sweden.

\bibitem[{Kusupati et~al.(2022)Kusupati, Bhatt, Rege, Wallingford, Sinha, Ramanujan, Howard-Snyder, Chen, Kakade, Jain, and Farhadi}]{NEURIPS2022_c32319f4}
Aditya Kusupati, Gantavya Bhatt, Aniket Rege, Matthew Wallingford, Aditya Sinha, Vivek Ramanujan, William Howard-Snyder, Kaifeng Chen, Sham Kakade, Prateek Jain, and Ali Farhadi. 2022.
\newblock \href {https://proceedings.neurips.cc/paper_files/paper/2022/file/c32319f4868da7613d78af9993100e42-Paper-Conference.pdf} {Matryoshka representation learning}.
\newblock In \emph{Advances in Neural Information Processing Systems}, volume~35, pages 30233--30249. Curran Associates, Inc.

\bibitem[{Liu and Xiong(2011)}]{liu2011survey}
Fasheng Liu and Lu~Xiong. 2011.
\newblock Survey on text clustering algorithm-research present situation of text clustering algorithm.
\newblock In \emph{2011 IEEE 2nd International Conference on Software Engineering and Service Science}, pages 196--199. IEEE.

\bibitem[{Mikolov et~al.(2013)Mikolov, Sutskever, Chen, Corrado, and Dean}]{mikolov2013distributed}
Tomas Mikolov, Ilya Sutskever, Kai Chen, Greg~S Corrado, and Jeff Dean. 2013.
\newblock Distributed representations of words and phrases and their compositionality.
\newblock \emph{Advances in neural information processing systems}, 26.

\bibitem[{Muennighoff(2022)}]{muennighoff2022sgpt}
Niklas Muennighoff. 2022.
\newblock Sgpt: Gpt sentence embeddings for semantic search.
\newblock \emph{arXiv preprint arXiv:2202.08904}.

\bibitem[{Muennighoff et~al.(2024)Muennighoff, Su, Wang, Yang, Wei, Yu, Singh, and Kiela}]{muennighoff2024generative}
Niklas Muennighoff, Hongjin Su, Liang Wang, Nan Yang, Furu Wei, Tao Yu, Amanpreet Singh, and Douwe Kiela. 2024.
\newblock Generative representational instruction tuning.
\newblock \emph{arXiv preprint arXiv:2402.09906}.

\bibitem[{Muennighoff et~al.(2023)Muennighoff, Tazi, Magne, and Reimers}]{muennighoff-etal-2023-mteb}
Niklas Muennighoff, Nouamane Tazi, Loic Magne, and Nils Reimers. 2023.
\newblock \href {https://doi.org/10.18653/v1/2023.eacl-main.148} {{MTEB}: Massive text embedding benchmark}.
\newblock In \emph{Proceedings of the 17th Conference of the European Chapter of the Association for Computational Linguistics}, pages 2014--2037, Dubrovnik, Croatia. Association for Computational Linguistics.

\bibitem[{Navjord and Korsvik(2023)}]{navjord2023beyond}
J{\o}rgen~Johnsen Navjord and Jon-Mikkel~Ryen Korsvik. 2023.
\newblock Beyond extractive: advancing abstractive automatic text summarization in norwegian with transformers.
\newblock Master's thesis, Norwegian University of Life Sciences, {\AA}s.

\bibitem[{Ni et~al.(2022)Ni, Hernandez~Abrego, Constant, Ma, Hall, Cer, and Yang}]{ni2021sentence}
Jianmo Ni, Gustavo Hernandez~Abrego, Noah Constant, Ji~Ma, Keith Hall, Daniel Cer, and Yinfei Yang. 2022.
\newblock \href {https://doi.org/10.18653/v1/2022.findings-acl.146} {Sentence-t5: Scalable sentence encoders from pre-trained text-to-text models}.
\newblock In \emph{Findings of the Association for Computational Linguistics: ACL 2022}, pages 1864--1874, Dublin, Ireland. Association for Computational Linguistics.

\bibitem[{Nielsen(2023)}]{nielsen-2023-scandeval}
Dan Nielsen. 2023.
\newblock \href {https://aclanthology.org/2023.nodalida-1.20} {{S}cand{E}val: A benchmark for {S}candinavian natural language processing}.
\newblock In \emph{Proceedings of the 24th Nordic Conference on Computational Linguistics (NoDaLiDa)}, pages 185--201, T{\'o}rshavn, Faroe Islands. University of Tartu Library.

\bibitem[{Nielsen(2016)}]{lcc}
Finn~Årup Nielsen. 2016.
\newblock Lcc.
\newblock \url{https://github.com/fnielsen/lcc-sentiment}.

\bibitem[{Nivre et~al.(2017)Nivre, Zeman, Ginter, and Tyers}]{nivre-etal-2017-universal}
Joakim Nivre, Daniel Zeman, Filip Ginter, and Francis Tyers. 2017.
\newblock \href {https://aclanthology.org/E17-5001} {{U}niversal {D}ependencies}.
\newblock In \emph{Proceedings of the 15th Conference of the {E}uropean Chapter of the Association for Computational Linguistics: Tutorial Abstracts}, Valencia, Spain. Association for Computational Linguistics.

\bibitem[{N{\o}rregaard and Derczynski(2021)}]{norregaard-derczynski-2021-danfever}
Jeppe N{\o}rregaard and Leon Derczynski. 2021.
\newblock \href {https://aclanthology.org/2021.nodalida-main.47} {{D}an{FEVER}: claim verification dataset for {D}anish}.
\newblock In \emph{Proceedings of the 23rd Nordic Conference on Computational Linguistics (NoDaLiDa)}, pages 422--428, Reykjavik, Iceland (Online). Link{\"o}ping University Electronic Press, Sweden.

\bibitem[{Pauli et~al.(2021)Pauli, Barrett, Lacroix, and Hvingelby}]{pauli-etal-2021-danlp}
Amalie~Brogaard Pauli, Maria Barrett, Oph{\'e}lie Lacroix, and Rasmus Hvingelby. 2021.
\newblock \href {https://aclanthology.org/2021.nodalida-main.53} {{D}a{NLP}: An open-source toolkit for {D}anish natural language processing}.
\newblock In \emph{Proceedings of the 23rd Nordic Conference on Computational Linguistics (NoDaLiDa)}, pages 460--466, Reykjavik, Iceland (Online). Link{\"o}ping University Electronic Press, Sweden.

\bibitem[{Pennington et~al.(2014)Pennington, Socher, and Manning}]{pennington2014glove}
Jeffrey Pennington, Richard Socher, and Christopher~D Manning. 2014.
\newblock Glove: Global vectors for word representation.
\newblock In \emph{Proceedings of the 2014 conference on empirical methods in natural language processing (EMNLP)}, pages 1532--1543.

\bibitem[{Reimers and Gurevych(2019{\natexlab{a}})}]{reimers2019sentence}
Nils Reimers and Iryna Gurevych. 2019{\natexlab{a}}.
\newblock \href {https://api.semanticscholar.org/CorpusID:201646309} {Sentence-bert: Sentence embeddings using siamese bert-networks}.
\newblock In \emph{Conference on Empirical Methods in Natural Language Processing}.

\bibitem[{Reimers and Gurevych(2019{\natexlab{b}})}]{reimers-gurevych-2019-sentence}
Nils Reimers and Iryna Gurevych. 2019{\natexlab{b}}.
\newblock \href {https://doi.org/10.18653/v1/D19-1410} {Sentence-{BERT}: Sentence embeddings using {S}iamese {BERT}-networks}.
\newblock In \emph{Proceedings of the 2019 Conference on Empirical Methods in Natural Language Processing and the 9th International Joint Conference on Natural Language Processing (EMNLP-IJCNLP)}, pages 3982--3992, Hong Kong, China. Association for Computational Linguistics.

\bibitem[{Reimers and Gurevych(2020)}]{reimers-gurevych-2020-making}
Nils Reimers and Iryna Gurevych. 2020.
\newblock \href {https://doi.org/10.18653/v1/2020.emnlp-main.365} {Making monolingual sentence embeddings multilingual using knowledge distillation}.
\newblock In \emph{Proceedings of the 2020 Conference on Empirical Methods in Natural Language Processing (EMNLP)}, pages 4512--4525, Online. Association for Computational Linguistics.

\bibitem[{Rekathati(2021)}]{rekathati2021introducing}
Faton Rekathati. 2021.
\newblock \href {https://kb-labb.github.io/posts/2021-08-23-a-swedish-sentence-transformer/} {The kblab blog: Introducing a swedish sentence transformer}.

\bibitem[{Samuel et~al.(2023)Samuel, Kutuzov, Touileb, Velldal, {\O}vrelid, R{\o}nningstad, Sigdel, and Palatkina}]{samuel-etal-2023-norbench}
David Samuel, Andrey Kutuzov, Samia Touileb, Erik Velldal, Lilja {\O}vrelid, Egil R{\o}nningstad, Elina Sigdel, and Anna Palatkina. 2023.
\newblock \href {https://aclanthology.org/2023.nodalida-1.61} {{N}or{B}ench {--} a benchmark for {N}orwegian language models}.
\newblock In \emph{Proceedings of the 24th Nordic Conference on Computational Linguistics (NoDaLiDa)}, pages 618--633, T{\'o}rshavn, Faroe Islands. University of Tartu Library.

\bibitem[{Sigurbergsson and Derczynski(2020)}]{sigurbergsson-derczynski-2020-offensive}
Gudbjartur~Ingi Sigurbergsson and Leon Derczynski. 2020.
\newblock \href {https://aclanthology.org/2020.lrec-1.430} {Offensive language and hate speech detection for {D}anish}.
\newblock In \emph{Proceedings of the Twelfth Language Resources and Evaluation Conference}, pages 3498--3508, Marseille, France. European Language Resources Association.

\bibitem[{Song et~al.(2020)Song, Tan, Qin, Lu, and Liu}]{song2020mpnet}
Kaitao Song, Xu~Tan, Tao Qin, Jianfeng Lu, and Tie-Yan Liu. 2020.
\newblock Mpnet: Masked and permuted pre-training for language understanding.
\newblock \emph{Advances in Neural Information Processing Systems}, 33:16857--16867.

\bibitem[{Str{\o}mberg-Derczynski et~al.(2021)Str{\o}mberg-Derczynski, Ciosici, Baglini, Christiansen, Dalsgaard, Fusaroli, Henrichsen, Hvingelby, Kirkedal, Kjeldsen, Ladefoged, Nielsen, Madsen, Petersen, Rystr{\o}m, and Varab}]{stromberg-derczynski-etal-2021-danish}
Leon Str{\o}mberg-Derczynski, Manuel Ciosici, Rebekah Baglini, Morten~H. Christiansen, Jacob~Aarup Dalsgaard, Riccardo Fusaroli, Peter~Juel Henrichsen, Rasmus Hvingelby, Andreas Kirkedal, Alex~Speed Kjeldsen, Claus Ladefoged, Finn~{\AA}rup Nielsen, Jens Madsen, Malte~Lau Petersen, Jonathan~Hvithamar Rystr{\o}m, and Daniel Varab. 2021.
\newblock \href {https://aclanthology.org/2021.nodalida-main.46} {The {D}anish {G}igaword corpus}.
\newblock In \emph{Proceedings of the 23rd Nordic Conference on Computational Linguistics (NoDaLiDa)}, pages 413--421, Reykjavik, Iceland (Online). Link{\"o}ping University Electronic Press, Sweden.

\bibitem[{Su et~al.(2020)Su, Zhang, and Immel}]{su2020digital}
Fangli Su, Yin Zhang, and Zachary Immel. 2020.
\newblock Digital humanities research: interdisciplinary collaborations, themes and implications to library and information science.
\newblock \emph{Journal of Documentation}, 77(1):143--161.

\bibitem[{Su et~al.(2023)Su, Shi, Kasai, Wang, Hu, Ostendorf, Yih, Smith, Zettlemoyer, and Yu}]{su2022one}
Hongjin Su, Weijia Shi, Jungo Kasai, Yizhong Wang, Yushi Hu, Mari Ostendorf, Wen-tau Yih, Noah~A. Smith, Luke Zettlemoyer, and Tao Yu. 2023.
\newblock \href {https://doi.org/10.18653/v1/2023.findings-acl.71} {One embedder, any task: Instruction-finetuned text embeddings}.
\newblock In \emph{Findings of the Association for Computational Linguistics: ACL 2023}, pages 1102--1121, Toronto, Canada. Association for Computational Linguistics.

\bibitem[{{Tatoeba Project Contributors}(2023)}]{TatoebaCorpus}
{Tatoeba Project Contributors}. 2023.
\newblock {Tatoeba Corpus}.
\newblock \url{https://tatoeba.org/}.
\newblock Used the version available at https://github.com/facebookresearch/LASER/tree/main/data/tatoeba/v1.

\bibitem[{Thakur et~al.(2021)Thakur, Reimers, R\"{u}ckl\'{e}, Srivastava, and Gurevych}]{Thakur2021BEIRAH}
Nandan Thakur, Nils Reimers, Andreas R\"{u}ckl\'{e}, Abhishek Srivastava, and Iryna Gurevych. 2021.
\newblock \href {https://datasets-benchmarks-proceedings.neurips.cc/paper_files/paper/2021/file/65b9eea6e1cc6bb9f0cd2a47751a186f-Paper-round2.pdf} {Beir: A heterogeneous benchmark for zero-shot evaluation of information retrieval models}.
\newblock In \emph{Proceedings of the Neural Information Processing Systems Track on Datasets and Benchmarks}, volume~1. Curran.

\bibitem[{Tiedemann(2012)}]{TIEDEMANN12.463}
Jörg Tiedemann. 2012.
\newblock Parallel data, tools and interfaces in {OPUS}.
\newblock In \emph{Proceedings of the Eight International Conference on Language Resources and Evaluation (LREC'12)}, Istanbul, Turkey. European Language Resources Association (ELRA).

\bibitem[{Velldal et~al.(2018)Velldal, {\O}vrelid, Bergem, Stadsnes, Touileb, and J{\o}rgensen}]{velldal-etal-2018-norec}
Erik Velldal, Lilja {\O}vrelid, Eivind~Alexander Bergem, Cathrine Stadsnes, Samia Touileb, and Fredrik J{\o}rgensen. 2018.
\newblock \href {https://aclanthology.org/L18-1661} {{N}o{R}e{C}: The {N}orwegian review corpus}.
\newblock In \emph{Proceedings of the Eleventh International Conference on Language Resources and Evaluation ({LREC} 2018)}, Miyazaki, Japan. European Language Resources Association (ELRA).

\bibitem[{Wang et~al.(2019)Wang, Pruksachatkun, Nangia, Singh, Michael, Hill, Levy, and Bowman}]{wang2019superglue}
Alex Wang, Yada Pruksachatkun, Nikita Nangia, Amanpreet Singh, Julian Michael, Felix Hill, Omer Levy, and Samuel Bowman. 2019.
\newblock Superglue: A stickier benchmark for general-purpose language understanding systems.
\newblock \emph{Advances in neural information processing systems}, 32.

\bibitem[{Wang et~al.(2018)Wang, Singh, Michael, Hill, Levy, and Bowman}]{wang-etal-2018-glue}
Alex Wang, Amanpreet Singh, Julian Michael, Felix Hill, Omer Levy, and Samuel Bowman. 2018.
\newblock \href {https://doi.org/10.18653/v1/W18-5446} {{GLUE}: A multi-task benchmark and analysis platform for natural language understanding}.
\newblock In \emph{Proceedings of the 2018 {EMNLP} Workshop {B}lackbox{NLP}: Analyzing and Interpreting Neural Networks for {NLP}}, pages 353--355, Brussels, Belgium. Association for Computational Linguistics.

\bibitem[{Wang et~al.(2022)Wang, Yang, Huang, Jiao, Yang, Jiang, Majumder, and Wei}]{wang2022text}
Liang Wang, Nan Yang, Xiaolong Huang, Binxing Jiao, Linjun Yang, Daxin Jiang, Rangan Majumder, and Furu Wei. 2022.
\newblock Text embeddings by weakly-supervised contrastive pre-training.
\newblock \emph{arXiv preprint arXiv:2212.03533}.

\bibitem[{Wang et~al.(2023)Wang, Yang, Huang, Yang, Majumder, and Wei}]{wang2023improving}
Liang Wang, Nan Yang, Xiaolong Huang, Linjun Yang, Rangan Majumder, and Furu Wei. 2023.
\newblock Improving text embeddings with large language models.
\newblock \emph{arXiv preprint arXiv:2401.00368}.

\bibitem[{Wang et~al.(2021)Wang, Bao, Huang, Dong, and Wei}]{wang-etal-2021-minilmv2}
Wenhui Wang, Hangbo Bao, Shaohan Huang, Li~Dong, and Furu Wei. 2021.
\newblock \href {https://doi.org/10.18653/v1/2021.findings-acl.188} {{M}ini{LM}v2: Multi-head self-attention relation distillation for compressing pretrained transformers}.
\newblock In \emph{Findings of the Association for Computational Linguistics: ACL-IJCNLP 2021}, pages 2140--2151, Online. Association for Computational Linguistics.

\bibitem[{Xiao et~al.(2023)Xiao, Liu, Zhang, and Muennighoff}]{xiao2023c}
Shitao Xiao, Zheng Liu, Peitian Zhang, and Niklas Muennighoff. 2023.
\newblock C-pack: Packaged resources to advance general chinese embedding.
\newblock Under review at ACL ARR.

\bibitem[{Xue et~al.(2021)Xue, Constant, Roberts, Kale, Al-Rfou, Siddhant, Barua, and Raffel}]{xue2020mt5}
Linting Xue, Noah Constant, Adam Roberts, Mihir Kale, Rami Al-Rfou, Aditya Siddhant, Aditya Barua, and Colin Raffel. 2021.
\newblock \href {https://doi.org/10.18653/v1/2021.naacl-main.41} {m{T}5: A massively multilingual pre-trained text-to-text transformer}.
\newblock In \emph{Proceedings of the 2021 Conference of the North American Chapter of the Association for Computational Linguistics: Human Language Technologies}, pages 483--498, Online. Association for Computational Linguistics.

\bibitem[{Zhang et~al.(2023)Zhang, Thakur, Ogundepo, Kamalloo, Alfonso-Hermelo, Li, Liu, Rezagholizadeh, and Lin}]{zhang2023miracl}
Xinyu Zhang, Nandan Thakur, Odunayo Ogundepo, Ehsan Kamalloo, David Alfonso-Hermelo, Xiaoguang Li, Qun Liu, Mehdi Rezagholizadeh, and Jimmy Lin. 2023.
\newblock Miracl: A multilingual retrieval dataset covering 18 diverse languages.
\newblock \emph{Transactions of the Association for Computational Linguistics}, 11:1114--1131.

\end{thebibliography}
\bibliographystyle{acl_natbib}

\section*{Checklist}

%%% BEGIN INSTRUCTIONS %%%
The checklist follows the references.  Please
read the checklist guidelines carefully for information on how to answer these
questions.  For each question, change the default \answerTODO{} to \answerYes{},
\answerNo{}, or \answerNA{}.  You are strongly encouraged to include a {\bf
justification to your answer}, either by referencing the appropriate section of
your paper or providing a brief inline description.  For example:
\begin{itemize}
  \item Did you include the license to the code and datasets? \answerYes{See Section~\ref{gen_inst}.}
  \item Did you include the license to the code and datasets? \answerNo{The code and the data are proprietary.}
  \item Did you include the license to the code and datasets? \answerNA{}
\end{itemize}
Please do not modify the questions and only use the provided macros for your
answers.  Note that the Checklist section does not count towards the page
limit.  In your paper, please delete this instructions block and only keep the
Checklist section heading above along with the questions/answers below.
%%% END INSTRUCTIONS %%%

\begin{enumerate}

\item For all authors...
\begin{enumerate}
  \item Do the main claims made in the abstract and introduction accurately reflect the paper's contributions and scope?
    \answerYes{}
  \item Did you describe the limitations of your work?
    \answerYes{}
  \item Did you discuss any potential negative societal impacts of your work?
    \answerNA{}
  \item Have you read the ethics review guidelines and ensured that your paper conforms to them?
    \answerYes{}
\end{enumerate}

\item If you are including theoretical results...
\begin{enumerate}
  \item Did you state the full set of assumptions of all theoretical results?
    \answerNA{}
	\item Did you include complete proofs of all theoretical results?
    \answerNA{}
\end{enumerate}

\item If you ran experiments (e.g. for benchmarks)...
\begin{enumerate}
  \item Did you include the code, data, and instructions needed to reproduce the main experimental results (either in the supplemental material or as a URL)?
    \answerYes{Code and dashboard is available at \url{https://github.com/KennethEnevoldsen/scandinavian-embedding-benchmark/}}
  \item Did you specify all the training details (e.g., data splits, hyperparameters, how they were chosen)?
    \answerYes{Dataset construction is described in the appendix, and data splits are specified in the task implementation.}
	\item Did you report error bars (e.g., with respect to the random seed after running experiments multiple times)?
    \answerYes{We specify the confidence interval for the rank but not for the specific task.}
	\item Did you include the total amount of compute and the type of resources used (e.g., type of GPUs, internal cluster, or cloud provider)?
    \answerYes{Specified in acknowledgements}
\end{enumerate}

\item If you are using existing assets (e.g., code, data, models) or curating/releasing new assets...
\begin{enumerate}
  \item If your work uses existing assets, did you cite the creators?
    \answerYes{See \autoref{tab:dataset_desc} for a full overview.}
  \item Did you mention the license of the assets?
    \answerYes{An overview of dataset licenses is available at \url{https://kennethenevoldsen.github.io/scandinavian-embedding-benchmark/datasets/}}
  \item Did you include any new assets either in the supplemental material or as a URL?
    \answerYes{The benchmark includes a publicly available dashboard available at: \url{https://kennethenevoldsen.github.io/scandinavian-embedding-benchmark/}}
  \item Did you discuss whether and how consent was obtained from people whose data you're using/curating?
    \answerNA{}
  \item Did you discuss whether the data you are using/curating contains personally identifiable information or offensive content?
    \answerNo{The DKHate datasets contain offensive content, given the nature of the task. As far as the authors know, none of the other tasks contain offensive content. Where applicable, PII has been anonymized by dataset distributors.}
\end{enumerate}

\item If you used crowdsourcing or conducted research with human subjects...
\begin{enumerate}
  \item Did you include the full text of instructions given to participants and screenshots, if applicable?
    \answerNA{}
  \item Did you describe any potential participant risks, with links to Institutional Review Board (IRB) approvals, if applicable?
    \answerNA{}
  \item Did you include the estimated hourly wage paid to participants and the total amount spent on participant compensation?
    \answerNA{}
\end{enumerate}

\end{enumerate}

\appendix

\newpage
\section{Appendix}

\subsection{Models}
The \autoref{tab:model_overview} reference to each of the model's names denoted in the main paper, which have been shortened for clarity.

{
\footnotesize
\tabcolsep=2pt
\begin{table*}
    \centering

\caption{This table provides an overview, along with reference to their implementation. If a link isn't provided, it denotes the name on Huggingface.}
    \label{tab:model_overview}
    
    \begin{tabular}{l|l}
        \toprule

         Name & Reference \\
         \midrule
         \midrule
         \textit{Self-Supervised Models}  \\
         \midrule
            dfm-encoder-large           & \texttt{danish-foundation-models/encoder-large-v1} \\
            \quad + SimCSE              & \url{Anonymized} \\   
            xlm-roberta-large           & \texttt{FacebookAI/xlm-roberta-large} \\
            nb-bert-large               &  \texttt{NbAiLab/nb-bert-large} \\
            nb-bert-base                & \texttt{NbAiLab/nb-bert-base} \\
            bert-base-swedish           & \texttt{KBLab/bert-base-swedish-cased} \\
            fasttext-cc-da              & \url{https://fasttext.cc/docs/en/crawl-vectors.html} \\
            fasttext-cc-nn              & \url{https://fasttext.cc/docs/en/crawl-vectors.html} \\
            fasttext-cc-nb              & \url{https://fasttext.cc/docs/en/crawl-vectors.html} \\
            fasttext-cc-sv              & \url{https://fasttext.cc/docs/en/crawl-vectors.html}   \\
        \midrule
        \textit{Supervised Models} \\
        \midrule
            multilingual-MiniLM-L12     & \texttt{sentence-transformers/paraphrase-multilingual-MiniLM-L12-v2} \\
            multilingual-mpnet-base     & \texttt{sentence-transformers/paraphrase-multilingual-mpnet-base-v2}\\
            labSE                       & \texttt{sentence-transformers/LaBSE} \\
           sentence-bert-swedish        & \texttt{KBLab/sentence-bert-swedish-cased} \\
            e5-mistral-7b-instruct  & \texttt{intfloat/e5-mistral-7b-instruct} \\
            multilingual-e5-large       & \texttt{intfloat/multilingual-e5-large} \\
            multilingual-e5-base        &  \texttt{intfloat/multilingual-e5-base} \\
            multilingual-e5-small       &  \texttt{intfloat/multilingual-e5-small} \\
            translate-e5-large          & Custom Implementation  \\
            sonar-dan                   & \texttt{facebook/SONAR}  \\
            sonar-nob                   & \texttt{facebook/SONAR} \\
            sonar-nno                   & \texttt{facebook/SONAR}  \\
            sonar-swe                   & \texttt{facebook/SONAR}  \\
        \midrule
        \textit{Embedding APIs} \\
        \midrule
            text-embedding-3-large      & \url{https://openai.com/blog/new-embedding-models-and-api-updates} \\ 
            text-embedding-3-small      &  \url{https://openai.com/blog/new-embedding-models-and-api-updates}  \\
            embed-multilingual-v3.0     & \url{https://txt.cohere.com/introducing-embed-v3/} \\
         \toprule
    \end{tabular}
\end{table*}
}

\subsection{Domains Generalization}

We see the performance across domains in \autoref{tab:result_across_domain}. These results are generally in accordance with the results across tasks; showing that the e5 models along with the commercial APIs constitute the most competitive models.

{
\footnotesize
\tabcolsep=2pt
\begin{table*}
    \centering

    \caption{Performance across domains in \gls{seb}. The best score in each model category is highlighted in bold. We only include domains that contain at least two datasets. Additional model evaluation can be found on the public Dashboard.}
    
    \begin{tabular}{l|c|ccc c ccc ccc}
        \toprule
         & Avg. & Fiction & Legal & News & N.-fiction & Review & Social & Spoken & Web & Wiki  \\
         \midrule
         Num. Datasets ($\rightarrow$) & 24 & 4 & 2 & 6 & 13 & 2 & 6 & 4 & 3 & 6\\ 
         \midrule
         \midrule
         \textit{Self-Supervised Models}  \\
         \midrule
            dfm-encoder-large           & 41.4 & 44.5 & 69.7 & 31.4 & 33.6 & 56.8 & 42.3 & 57.0 & 29.4 & 31.0  \\
            \quad + SimCSE              & \textbf{46.6} & \textbf{46.4} & 72.0 & \textbf{40.5} & \textbf{42.7} & 58.7 & \textbf{41.2} & 60.7 &\textbf{ 39.3} & 37.3\\
            xlm-roberta-large           & 35.3 & 41.5 & 41.3 & 24.9 & 25.3 & 55.9 & 36.2 & 54.4 & 24.4 & 26.5 \\
            nb-bert-large               & 46.0 & 44.0 & \textbf{73.2} & 38.7 & 42.6 & \textbf{61.6} & 36.1   & \textbf{61.7} & 30.5 & \textbf{39.9}\\
            nb-bert-base                & 42.1  & 42.6 & 71.8 & 28.7 & 35.1 & 57.6 & 38.4 & 58.7 & 29.0 & 35.0 \\
            bert-base-swedish           & 35.2 & 38.6 & 56.4 & 24.9 & 29.9 & 56.9 & 29.8 & 49.7 & 27.3 & 25.0 \\
            fasttext-cc-da              & 37.3 & 39.5 & 64.3 & 28.4 & 34.0 & 49.9 & 33.2 & 45.6 & 26.0 & 33.9 \\
            fasttext-cc-nn              & 35.8  & 38.1 & 64.2 & 24.8 & 33.6 & 47.5 & 29.2 & 43.2 & 24.0 & 35.5 \\
            fasttext-cc-nb              & 37.5 & 39.0 & 63.5 & 26.8 & 34.4 & 49.8 & 32.0 & 46.1 & 25.4 & 36.5 \\
            fasttext-cc-sv              & 36.0 & 38.3 & 62.7 & 28.0 & 33.3 & 50.9 & 30.1 & 45.8 & 26.6 & 29.3   \\
        \midrule
        \textit{Supervised Models} \\
        \midrule
            multilingual-MiniLM-L12     & 50.0 & 43.5 & 68.4 & 43.9 & 49.1 & 59.9 & 45.4 & 57.6 & 43.6 & 41.2 \\
            multilingual-mpnet-base     & 53.2 & 44.2 & 72.8 & 47.3 & 52.4 & 64.7 & 49.0 & 59.7 & 45.6 & 43.3 \\
            labSE                       & 50.5 & 49.0 & 71.3 & 41.9 & 48.5 & 61.9 & 48.5 & 57.7 & 48.6 & 44.6 \\
           sentence-bert-swedish        & 46.6 & 40.4 & 59.9 & 44.1 & 47.1 & 57.5 & 36.8 & 53.9 & 44.9 & 36.1 \\
            e5-mistral-7b-instruct      & 60.4 & \textbf{53.7} & \textbf{77.6} & 52.3 & 58.0 & 70.1 & \textbf{58.0} & \textbf{64.5} & \textbf{62.1} & \textbf{57.0}    \\
            multilingual-e5-large       & \textbf{60.7} & 48.1 & 76.1 & \textbf{54.5} & \textbf{58.9} & \textbf{73.5} & 54.9 & 62.0 & 54.9 & 55.7 \\
            multilingual-e5-base        & 57.9 & 48.5 & 74.9 & 50.4 & 56.2 & 69.6 & 52.6 & 59.7 & 54.3 & 54.8 \\
            multilingual-e5-small       &  56.4 & 49.0 & 72.3 & 50.8 & 55.4 & 65.9 & 51.1 & 57.8 & 54.8 & 53.4  \\
            translate-e5-large          & 47.7 & 43.2 & 69.4 & 36.8 & 43.7 & 68.1 & 46.5 & 55.5 & 40.1 & 37.8    \\
            sonar-dan                   & 43.4 & 50.2 & 73.5 & 31.0 & 35.7 & 59.1 & 49.2 & 55.5 & 43.0 & 33.1  \\
            sonar-nob                   & 41.5 & 45.2 & 70.1 & 28.0 & 34.1 & 57.9 & 43.8 & 55.6 & 35.8 & 31.0 \\
            sonar-nno                   & 41.5 & 46.5 & 71.3 & 28.4 & 33.9 & 58.5 & 44.8 & 56.0 & 37.7 & 30.0  \\
            sonar-swe                   & 42.8 & 50.5 & 73.2 & 30.9 & 35.9 & 58.2 & 47.0 & 55.0 & 44.1 & 33.5  \\
        \midrule
        \textit{Embedding APIs} \\
        \midrule
            text-embedding-3-large      & \textbf{65.0} & \textbf{50.5} & 76.1 & 56.1 & \textbf{64.1} & 72.7 & \textbf{59.0} & \textbf{64.4} & \textbf{61.0} & \textbf{65.5} \\ 
            text-embedding-3-small      & 61.0 & 50.2 & 75.9 & 54.0 & 61.2 & 66.6 & 55.3 & 61.2 & 58.1 & 60.7 \\
            embed-multilingual-v3.0     & 64.1 & 49.2 & \textbf{76.6} & \textbf{56.2} & 63.5 & \textbf{75.2} & 57.1 & 63.3 & 57.9 & 63.6 \\
         \toprule
    \end{tabular}
    \label{tab:result_across_domain}
\end{table*}
}

\subsection{Dataset Embeddings}
\label{sec:appendix_embedding_map}

\begin{figure*}
    \centering
    \includegraphics[width=0.9\linewidth]{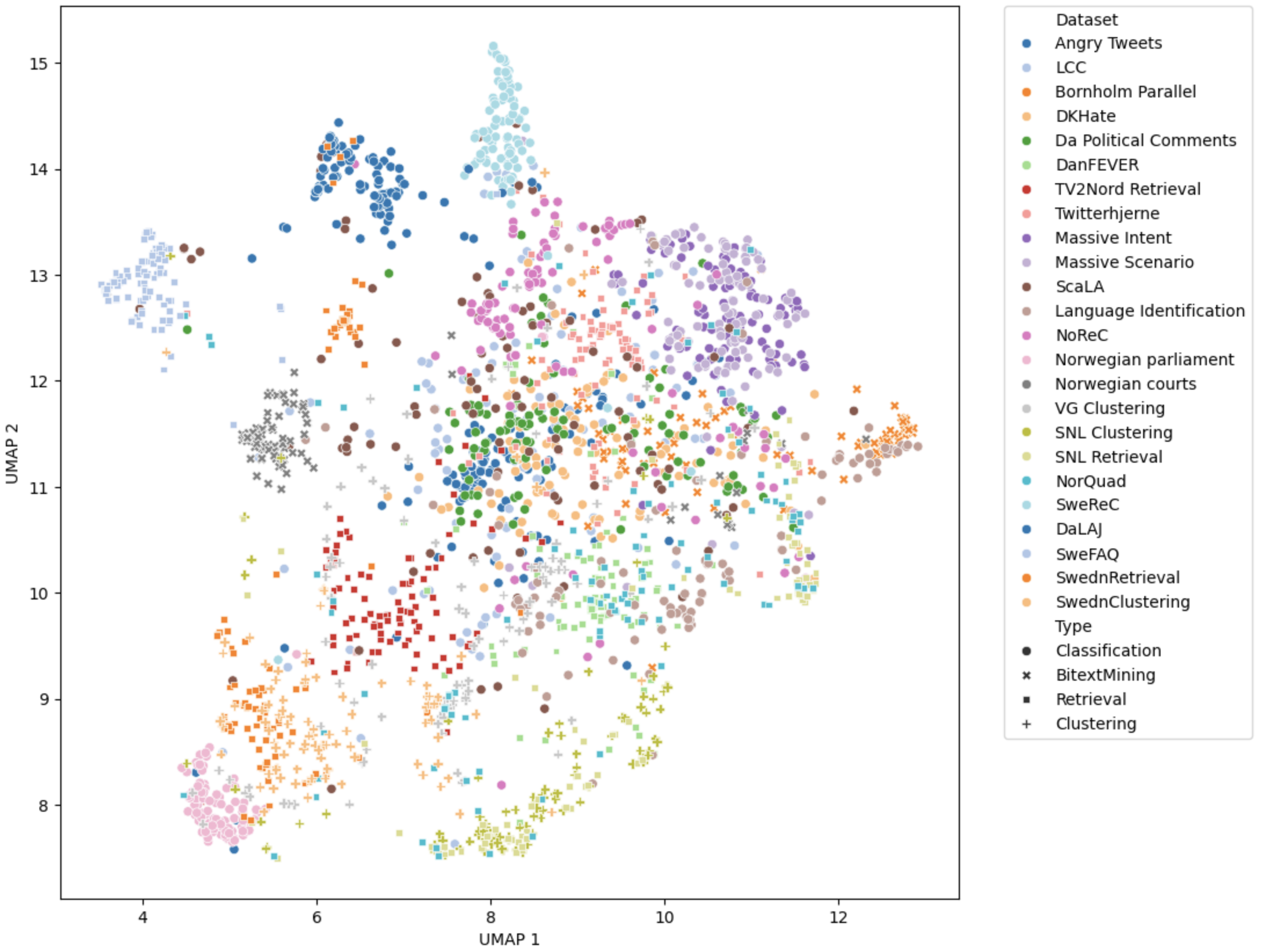}
    \caption{The embeddings of 100 randomly sampled documents from each task, embedded using embed-multilingual-v3.0 and projected using a UMAP projection. The project uses the cosine metrics but otherwise default parameter values.}
    \label{fig:emb_map}
\end{figure*}

To examine the spread and similarity of our datasets, we explore their similarity in the embedding space in \autoref{fig:emb_map}. To do so, we use one of the best-performing embedding models, embed-multilingual-v3.0. We see that certain datasets occupy distinct clusters, indicating that evaluations without these datasets would likely bias the model evaluation. Notably, we additionally see that the existing (translated) datasets within \gls{mteb} (Massive Intent and Massive Scenario) cover only a small subsection of the embedding space.

\subsection{Evaluation and Metrics}
\label{sec:app_evaluation}
This section briefly presents the tasks, their evaluation, and their metric. However, we utilize a similar implementation as \gls{mteb} to keep results comparable. Thus we refer to the original work for a more detailed introduction. We do, however, make one notable difference, that is, we allow the models to embed the tasks differently depending on the task, this is especially relevant for the e5 models, embed-multilingual-v3.0 and prompt-based models such as e5-mistral-7b-instruct.

\noindent
\textbf{Classification:}
Using the embedding model a train and a test set are embedded. Using the embedding training set a logistic classifier is trained using a maximum of 100 iterations. The model is then tested on the test set. This approach is repeated 10 times and metrics are calcuated for each set and aggregated. The training sets for these repeats are obtaining by sampling from the training set (with replacement) 16 examples from each label. The mean accuracy is reported as the main metric. Other metrics reported include the F1-score and  measures of uncertainty of metrics (standard error).

\noindent
\textbf{Bitext Mining:}
The dataset consists of matching pairs of sentences, and the goal is to find the match. All matching pairs of sentences are embedded using the embedding model. Afterward, the closest match is found using cosine similarity. F1 is reported as the main metric.

\noindent
\textbf{Clustering}
The dataset consists of documents attached with a label, e.g., a denoted category such as "sports." The goal is the correctly cluster the documents into similar clusters as the labels.  
All documents are embedded, and a mini-batch k-means model (batch size 32 and k equal to the number of unique labels in the dataset) is trained on the embeddings. The V measure is used as is reported as the main metric, as the permutation of labels does not affect the score.

\noindent
\textbf{Retrieval:}
The dataset consists of a corpus, queries as well as a mapping between the queries and their relevant documents. The goal is to retrieve these relevant documents. Both queries and documents are embedded using the model. We allow these to be embedded differently depending on the model. For each query, the corpus documents are ranked using a similarity score, and nDCG@10 is reported as the main metric.

\subsection{Datasets Construction}
\label{sec:app_dataset_construction}

This section briefly describes the construction of the tasks.

\noindent
\textbf{Classification:} As all the classification datasets are derived from existing datasets, no additional processing is done to these except to limit the size of excessively large datasets.

\noindent
\textbf{Bitext Mining:} Similar to the classification, these datasets already existed as paired datasets. With the Norwegian Courts being extracted from OPUS \citep{TIEDEMANN12.463} and Bornholm Parallel being derived from \citep{derczynski-kjeldsen-2019-bornholmsk}.

\noindent
\textbf{Clustering:} For clustering, we construct the dataset based on existing datasets of news or encyclopedic corpora \citep{navjord2023beyond, berdicevskis-etal-2023-superlim} using their attached categories. The SNL and VG datasets \citep{navjord2023beyond} contain a hierarchy of labels; here, we subjectively chose a meaning level and validated that it led to a meaningful separation in performance -- using either too many or too few levels would to either 1-2 clusters or clusters consisting of only 2-3 documents. 

Similar to the classification, these datasets already existed as paired datasets. With the Norwegian Courts being extracted from OPUS \citep{TIEDEMANN12.463} and Bornholm Parallel being derived from \citep{derczynski-kjeldsen-2019-bornholmsk}.

\noindent
\textbf{Retrieval:} 
For the construction of the retrieval datasets, we used either question and answer datasets (e.g., NorQuad \citep{ivanova-etal-2023-norquad}) or news summarization datasets (e.g., \citep{berdicevskis-etal-2023-superlim}). For the question and answer we specified the questions as queries and the answers as the corpus. A correct answer was considered to be a properly retrieved document. For the summaries, we considered the headline as the query and both the summaries and the articles as the corpus. Matching summaries and articles were considered properly retrieved documents. 

\subsection{Datasets Statistics}
\autoref{tab:dataset_desc} contains an overview of each of the datasets in \gls{seb}, including a short description, descriptive statics, task formalization, and domains as defined by \citep{nivre-etal-2017-universal}. 

%\begin{table}[h]
%\centering
\onecolumn
{\small
\setlength{\extrarowheight}{7pt}
\begin{longtable}{L{2.2cm}|L{2.8cm}L{1.4cm}L{1.0cm}lL{1.3cm}cp{1.3cm}}
\toprule
        \textbf{Dataset} & \textbf{Description} & \textbf{Main Score} & \textbf{Langs} & \textbf{Type} & \textbf{Domains} & \textbf{N. Docs} & \textbf{Avg. Length} \\ 
        \midrule
        \endhead
        Angry Tweets \cite{pauli-etal-2021-danlp} & A sentiment dataset with 3 classes (positiv, negativ, neutral) for Danish tweets & Accuracy & da & Classification & social & 1047 & 156.15 (82.02) \\
        Bornholm Parallel \cite{derczynski-kjeldsen-2019-bornholmsk} & Danish Bornholmsk Parallel Corpus. Bornholmsk is a Danish dialect spoken on the island of Bornholm, Denmark. & F1 & da, da-bornholm & BitextMining & poetry, wiki, fiction, web, social & 1000 & 44.36 (41.22) \\
        DKHate \cite{sigurbergsson-derczynski-2020-offensive} & Danish Tweets annotated for Hate Speech either being Offensive or not & Accuracy & da & Classification & social & 329 & 88.18 (68.30) \\ 
        Da Political Comments & A dataset of Danish political comments rated for sentiment & Accuracy & da & Classification & social & 7206 & 69.60 (62.85) \\
        DaLAJ \cite{berdicevskis-etal-2023-superlim} & A Swedish dataset for linguistic acceptability. Available as a part of Superlim & Accuracy & sv & Classification & fiction, non-fiction & 888 & 120.77 (67.95) \\
        DanFEVER \cite{norregaard-derczynski-2021-danfever} & A Danish dataset intended for misinformation research. It follows the same format as the English FEVER dataset. & NDCG@10 & da & Retrieval & wiki, non-fiction & 8897 & 124.84 (168.53) \\
        LCC \cite{lcc} & The Leipzig corpora collection, annotated for sentiment & Accuracy & da & Classification & legal, web, news, social, fiction, non-fiction, academic, government & 150 & 118.73 (57.82) \\ 
        Language Identification \cite{haas-derczynski-2021-discriminating} & A dataset for Nordic language identification. & Accuracy & da, sv, nb, nn, is, fo & Classification & wiki & 3000 & 78.23 (48.54) \\
        Massive Intent \cite{fitzgerald2022massive}  & The intent task within MASSIVE corpus translated for Scandinavian languages & Accuracy & da, nb, sv & Classification & spoken & 15021 & 34.65 (16.99) \\
        Massive Scenario \cite{fitzgerald2022massive} & The scenario task within MASSIVE corpus translated for Scandinavian languages & Accuracy & da, nb, sv & Classification & spoken & 15021 & 34.65 (16.99) \\
        NoReC \cite{velldal-etal-2018-norec} & A Norwegian dataset for sentiment classification on review & Accuracy & nb & Classification & reviews & 2048 & 89.62 (61.21) \\ 
        NorQuad \cite{ivanova-etal-2023-norquad} & Human-created question for Norwegian Wikipedia passages. & NDCG@10 & nb & Retrieval & non-fiction, wiki & 2602 & 502.19 (875.23) \\
        Norwegian courts \cite{TIEDEMANN12.463} & Nynorsk and Bokmål parallel corpus from Norwegian courts. & F1 & nb, nn & BitextMining & legal, non-fiction & 456 & 82.11 (49.48) \\
        Norwegian parliament & Norwegian parliament speeches annotated with the party of the speaker (`Sosialistisk Venstreparti` vs `Fremskrittspartiet`) & Accuracy & nb & Classification & spoken & 2400 & 1897.51 (1988.62) \\ 
        SNL Clustering \cite{navjord2023beyond} & Webscrabed articles from the Norwegian lexicon 'Det Store Norske Leksikon'. Uses article's categories as clusters. & V measure & nb & Clustering & non-fiction, wiki & 2048 & 1101.30 (2168.35) \\
        SNL Retrieval \cite{navjord2023beyond} & Webscrabed articles and ingresses from the Norwegian lexicon 'Det Store Norske Leksikon'. & NDCG@10 & nb & Retrieval & non-fiction, wiki & 2600 & 1001.43 (2537.83) \\
        ScaLA \cite{nielsen-2023-scandeval} & A linguistic acceptability task for Danish, Norwegian Bokmål Norwegian Nynorsk and Swedish. & Accuracy & da, nb, sv, nn & Classification & fiction, news, non-fiction, spoken, blog & 8192 & 102.45 (55.49) \\ 
        SweFAQ \cite{berdicevskis-etal-2023-superlim} & A Swedish QA dataset derived from FAQ & NDCG@10 & sv & Retrieval & non-fiction, web & 1024 & 195.44 (209.33) \\
        SweReC \cite{nielsen-2023-scandeval} & A Swedish dataset for sentiment classification on review & Accuracy & sv & Classification & reviews & 2048 & 318.83 (499.57) \\ 
        SwednClustering \cite{berdicevskis-etal-2023-superlim} & News articles from the Swedish newspaper Dagens Nyheter (DN) collected during the years 2000--2020. Uses the category labels as clusters. & V measure & sv & Clustering & non-fiction, news & 2048 & 1619.71 (2220.36) \\
        SwednRetrieval \cite{berdicevskis-etal-2023-superlim} & News articles from the Swedish newspaper Dagens Nyheter (DN) collected during the years 2000--2020. & NDCG@10 & sv & Retrieval & non-fiction, news & 3070 & 1946.35 (3071.98) \\ 
        TV2Nord Retrieval & News Article and corresponding summaries extracted from the Danish newspaper TV2 Nord. & NDCG@10 & da & Retrieval & news, non-fiction & 4096 & 784.11 (982.97) \\
        Twitterhjerne \cite{holm2024gllms} & Danish question asked on Twitter with the Hashtag \#Twitterhjerne ('Twitter brain') and their corresponding answer. & NDCG@10 & da & Retrieval & social & 340 & 138.23 (82.41) \\ 
        VG Clustering \cite{navjord2023beyond} & Articles and their classes (e.g. sports) from VG news articles extracted from Norsk Aviskorpus. & V measure & nb & Clustering & non-fiction, news & 2048 & 1009.65 (1597.60) \\ 
        \toprule
\caption{Tasks available in SEB. The average length is specified in characters. Values in parentheses denote the standard deviation.}
\label{tab:dataset_desc}
\end{longtable}
}
\clearpage

% use for ACL ARR
%\twocolumn

\subsection{Long-term Availability and Stability}
All of the datasets used for the Scandinavian Embedding Benchmark are publicly available on Huggingface repositories. To avoid duplicating metadata, we refer to and download from existing repositories, however, to ensure stability, we refer to a specific revision used. This allows us to update the benchmark datasets if annotations are corrected or faulty entries are removed. Additionally, we keep a copy of all datasets in case datasets are removed from the Huggingface Hub such that they can be re-uploaded. The permissible licenses of the datasets ensure that this is a viable option.

\subsection{Results per Task}
\label{sec:result_pr_task}

In the following figure, we see an overview of all of the results of the benchmark for each task for the selected models. To get an up-to-date overview, check out the online dashboard.

\end{document}